\documentclass{article}
\usepackage{iclr2021_conference,times}
\iclrfinalcopy
\usepackage{algorithm}
\usepackage{algorithmic}
\usepackage[utf8]{inputenc} % allow utf-8 input
\usepackage[T1]{fontenc}    % use 8-bit T1 fonts
\usepackage{hyperref}       % hyperlinks
\usepackage{url}            % simple URL typesetting
\usepackage{booktabs}       % professional-quality tables
\usepackage{amsfonts}       % blackboard math symbols
\usepackage{nicefrac}       % compact symbols for 1/2, etc.
\usepackage{microtype}      % microtypography
\usepackage{amsmath}
\usepackage{color}
\usepackage{amssymb}        % checkmark
\usepackage{caption} 
\usepackage{wrapfig}
\usepackage[pdftex]{graphicx}
\usepackage{subcaption}
\usepackage{hyperref}
\usepackage{amsmath}
\usepackage[title]{appendix}
\usepackage{mathtools}
\usepackage{pgf, tikz}
\usepackage{slashbox}
\usepackage{multirow}
\usepackage{junstyle}
\DeclarePairedDelimiterX{\infdivx}[2]{(}{)}{%
  #1\;\delimsize\|\;#2%
}

\newcommand{\distas}[1]{\mathbin{\overset{#1}{\kern\z@\sim}}}%

\usepackage{animate}
\newcommand{\comment}[1]{}

\usepackage{hyperref}
\usepackage{url}

\newtheorem{theorem}{Theorem}

\title{Disentangled Recurrent Wasserstein Autoencoder}

\author{Jun Han \thanks{Equal contribution.} \thanks{Part of this work was done before joining Tencent.}\\
PCG, Tencent \\
\texttt{junhanjh@tencent.com} \\
\hspace{-2.5cm}
\And
Martin Renqiang Min \footnotemark[1]\\
NEC Laboratories America \\
\texttt{renqiang@nec-labs.com} \hspace{-2.5cm} \\
\hspace{-2.5cm}
\And
Ligong Han \footnotemark[1] \\
Rutgers University \hspace{-2.5cm}\\
\texttt{hanligong@gmail.com} \\
\hspace{-2.5cm}
\And
Li Erran Li \\
Columbia University \\
\texttt{erranlli@gmail.com} 
\hspace{-2.5cm}
\And
Xuan Zhang \thanks{Work was done before joining Texas A\&M University.}\\
Texas A\&M University \\
\texttt{floatlazer@gmail.com} 
}

\begin{document}
\maketitle 

\begin{abstract}
Learning disentangled representations leads to interpretable models and facilitates data generation with style transfer, which has been extensively studied on static data such as images in an unsupervised learning framework. However, only a few works have explored unsupervised disentangled sequential representation learning due to challenges of generating sequential data. In this paper, we propose recurrent Wasserstein Autoencoder (R-WAE), a new framework for generative modeling of sequential data. R-WAE disentangles the representation of an input sequence into static and dynamic factors (i.e., time-invariant and time-varying parts). Our theoretical analysis shows that, R-WAE minimizes an upper bound of a penalized form of the Wasserstein distance between model distribution and sequential data distribution, and simultaneously maximizes the mutual information between input data and different disentangled latent factors, respectively. This is superior to (recurrent) VAE which does not explicitly enforce mutual information maximization between input data and disentangled latent representations. When the number of actions in sequential data is available as weak supervision information, R-WAE is extended to learn a categorical latent representation of actions to improve its disentanglement. Experiments on a variety of datasets show that our models outperform other baselines with the same settings in terms of disentanglement and unconditional video generation both quantitatively and qualitatively.
\end{abstract}

\section{Introduction}

Unsupervised representation learning is an important research topic in machine learning. It embeds high-dimensional sensory data such as images and videos into a low-dimensional latent space in an unsupervised learning framework, aiming at extracting essential data variation factors to help downstream tasks such as classification and prediction~\citep{bengio2013representation}. In the last several years, disentangled representation learning, which further separates the latent embedding space into exclusive explainable factors such that each factor only interprets one of semantic attributes of sensory data, has received a lot of interest and achieved many empirical successes on static data such as images~\citep{chen2016infogan, higgins2017beta, dupont2018learning, chen2018isolating, rubenstein2018learning, rubenstein2018latent, kim2018disentangling}. For example, the latent representation of handwritten digits can be disentangled into a content factor encoding digit identity and a style factor encoding handwriting style.

In spite of successes on static data, only a few works have explored unsupervised representation disentanglement of sequential data due to the challenges of developing generative models of sequential data. Learning disentangled representations of sequential data is important and has many applications. For example, the latent representation of a smiling-face video can be disentangled into a static part encoding the identity of the person (content factor) and a dynamic part encoding the smiling motion of the face (motion factor). The disentangled representation of the video can be potentially used for many downstream tasks such as classification, retrieval, and synthetic video generation with style transfer. Most of previous unsupervised representation disentanglement models for static data heavily rely on the KL-divergence regularization in a VAE framework~\citep{higgins2017beta, dupont2018learning, chen2018isolating, kim2018disentangling}, which has been shown to be problematic due to matching individual instead of aggregated posterior distribution of the latent code to the same prior~\citep{tolstikhin2017wasserstein, rubenstein2018learning, rubenstein2018latent}. Therefore, extending VAE or recurrent VAE~\citep{vrnn_nips2015} to disentangle sequential data in a generative model framework~\citep{hsu2017unsupervised,li2018disentangled} is not ideal. In addition, recent research~\citep{locatello2019challenging} has theoretically shown that it is impossible to perform unsupervised disentangled representation learning without inductive biases on both models and data, especially on static data. Fortunately, sequential data such as videos often have clear inductive biases for the disentanglement of content factor and motion factor as mentioned in~\citep{locatello2019challenging}. Unlike static data, the learned static and dynamic factors of sequential data are not exchangeable.
 
In this paper, we propose a recurrent Wasserstein Autoencoder (R-WAE) to learn disentangled representations of sequential data. We employ a Wasserstein metric~\citep{arjovsky2017wasserstein, gulrajani2017improved, bellemare2017cramer} induced from the optimal transport between model distribution and the underlying data distribution, which has some nicer properties (for e.g., sum invariance, scale sensitivity, applicable to distributions with non-overlapping supports, and better out-of-sample performance in the worst-case expectation~\citep{esfahani2018data}) than the  $\KL$ divergence in VAE~\citep{kingma2013auto} and $\beta$-VAE~\citep{higgins2017beta}. Leveraging explicit inductive biases in both sequential data and model, we encode an input sequence into two parts: a shared static latent code and a dynamic latent code, and sequentially decode each element of the sequence by combining both codes. We enforce a fixed prior distribution for the static code and learn a prior for the dynamic code to ensure the consistency of the sequence. The disentangled representations are learned by separately regularizing the posteriors of the latent codes with their corresponding priors. 

Our main contributions are summarized as follows: (1) We draw the first connection between minimizing a Wasserstein distance and maximizing mutual information for unsupervised representation disentanglement of sequential data from an information theory perspective; (2) We propose two sets of effective regularizers to learn the disentangled representation in a completely unsupervised manner with explicit inductive biases in both sequential data and models. %Empirical results show that both methods can exactly disentangle static and dynamic parts in a completely unsupervised manner and generate consistent long sequences of high quality on benchmark datasets. 
(3) We incorporate a relaxed discrete latent variable to improve the disentangled learning of actions on real data. Experiments show that our models achieve state-of-the-art performance in both disentanglement of static and dynamic latent representations and unconditional video generation under the same settings as baselines~\citep{li2018disentangled, tulyakov2018mocogan}.
%Our paper is organized as follows. Section 2 discusses related works. Section 3 introduces our main framework. Section 4 provides empirical results. We conclude the paper in Section 5.
% this is an unimportant paragraph due to limited space, and it's uncommon to put it in the introduction after summary.

%erran: related work, not related works
%\section{Notations and Related Works}
\section{Background and Related Work}
\paragraph{Notation} 
Let calligraphic letters (i.e. $\CX$) be sets, capital letters (i.e. $X$) be random variables and lowercase letters be their values. Let $\BD(P_X, P_G)$ be the divergence between the true (but unknown) data distribution $P_X$ (density p(x)) and the latent-variable generative model distribution $P_G$ specified by a prior distribution $P_Z$ (density p(z)) of latent variable $Z$. Let $\BD_{\KL}$ be $\KL$ divergence, $\BD_{\JS}$ be Jensen-Shannon divergence and $\MMD$ be Maximum Mean Discrepancy (MMD)~\citep{gretton2007kernel}. 

\paragraph{Optimal Transport Between Distributions} The optimal transport cost inducing a rich class of divergence between the distribution $P_X$ and the distribution $P_G$ is defined as follows,
\begin{equation}
W(P_X, P_G)\!\! :=\!\! \inf_{\Gamma \sim \cp(X\sim P_X, Y \sim P_G)} \!\!\!\!\! \E_{(X, Y)\sim \Gamma}[c(X, Y)],      
\end{equation}
where $c(X, Y)$ is any measurable cost function and $\cp(X\sim P_X, Y \sim P_G)$ is the set of joint distributions of (X, Y) with respective marginals $P_X$ and $P_G$. 

\paragraph{Comparison between WAE~\citep{tolstikhin2017wasserstein} and VAE~\citep{kingma2013auto}} Instead of optimizing over all couplings $\Gamma$ between two random variables in $\CX$,  \cite{bousquet2017optimal, tolstikhin2017wasserstein} show that it is sufficient to find $Q(Z|X)$ such that the marginal $Q(Z):=E_{X\sim P_X}[Q(Z|X)]$ is identical to the prior $P(Z),$  as given in the following definition,
\begin{mydef}
For any deterministic $P_G(X|Z)$ and any function $G:\CZ\rightarrow\CX,$
\begin{equation}
W(P_X, P_G)=\!\! \inf_{Q: Q_Z=P_Z} \!\! \E_{P_X}\E_{Q(Z\mid X)} [c(X, G(Z))].    
\end{equation}
\label{lem:wae}
\end{mydef}
\vspace{-0.2cm}
Definition~\ref{lem:wae} leads to the following loss $\BD_{\mathrm{WAE}}$ of WAE based on a Wasserstein distance, 
\begin{equation}
\inf_{Q(Z|X)}\E_{P_X}\E_{Q(Z\mid X)} [c(X, G(Z))]+\beta~\BD(Q_Z, P_Z),    
\end{equation}
where the first term is data reconstruction loss, and the second one is a regularizer that forces the posterior $Q_Z=\int Q(Z|X)d P_X$ to match the prior $P_Z$ (Adversarial autoencoder (AAE)~\citep{makhzani2015adversarial} shares a similar idea to WAE). In contrast, VAE has a different regularizer $\E_X[\BD_{\KL}(Q(Z|X), P_Z))]$ enforcing the latent posterior distribution of each input to match $P_Z$. In \citep{rubenstein2018latent, rubenstein2018learning}, it is shown that WAE has better disentanglement than $\beta$-VAE~\citep{higgins2017beta} on images, which inspires us to design a new representation disentanglement framework for sequential data with several innovations. %Although our method is inspired by WAE, there are several innovations to achieve our goal.  

\paragraph{Unsupervised disentangled representation learning} Several generative models have been proposed to learn disentangled representations of sequential data~\citep{denton2017unsupervised, hsu2017unsupervised, li2018disentangled, hsieh2018learning, sun2018two, tulyakov2018mocogan}. FHVAE in \citep{hsu2017unsupervised} is a VAE-based hierarchical graphical model with factorized Gaussian priors and only focuses on speech or audio data. Our R-WAE employing a more powerful recurrent prior can be applied to both speech and video data. The models in \citep{sun2018two, denton2017unsupervised, hsieh2018learning} are based on the first several elements of a sequence to design disentanglement architectures for future sequence predictions. %\cite{hsieh2018learning} also extends VAE and disentangles factors of objects in a sequence but only focuses on future frame prediction on toy datasets.

In terms of representation learning by mutual information maximization, our work empirically demonstrates that explicit inductive biases in data and model architecture are necessary to the success of learning meaningful disentangled representations of sequential data, while the works in~\citep{locatello2019challenging, poole2019variational, tschannen2019mutual, ozair2019wasserstein} are about general representation learning, especially on static data.

The most related works to ours are MoCoGAN~\citep{tulyakov2018mocogan} and DS-VAE~\citep{li2018disentangled},  which have the ability to disentangle variant and invariant parts of sequential data and perform  unconditional sequence generation. \cite{tulyakov2018mocogan} is a GAN-based model that can be only applied to the setting in which the number of motions is finite, and cannot encode the latent representation of sequences. \cite{li2018disentangled} provides a disentangled sequential autoencoder based on VAE~\citep{kingma2013auto}. Training VAE is equivalent to minimizing a lower bound of the KL divergence between empirical data distribution and generated data distribution, which has been shown to produce inferior disentangled representations of static data than generative models employing the Wasserstein metric~\citep{rubenstein2018latent, rubenstein2018learning}. %In addition, compared to our framework, \cite{li2018disentangled} has simpler generative and inference model architecture, which is less expressive for modeling complex sequential data. 

% Hi Jun, you want to say Yingzhen's work has simpler encoder/decoder architecture? But it's really not that big a deal

% we need to cite https://arxiv.org/pdf/1802.03761.pdf
% on the latent space of wasserstein autoencoders
%\textbf{erran: this criticism seems to be too strong given we use same datasets.}
\section{Proposed Approach: Disentangled Recurrent Wasserstein Autoencoder (R-WAE)}
\label{mainmethod}
% \begin{figure*}[ht]
% \centering
% \begin{tabular}{cccc}\hspace{-0.2cm}
% \includegraphics[height=0.166\textwidth]{figures/Gen.pdf}& \hspace{-0.55cm}
% \includegraphics[height=0.166\textwidth]{figures/Infer.pdf}& \hspace{-0.55cm}
% \includegraphics[height=0.166\textwidth]{figures/Regen.pdf}& \hspace{-0.55cm}
% \includegraphics[height=0.166\textwidth]{figures/Reinfer.pdf}\\
% \vspace{-0.2cm}
% {\small (a) Generative Model} & {\small\! (b) Inference Model} &  {\small \!\! \! \!\! (c) Weakly Supervised Generative}& {\small \!\!\!\!\! (d) Weakly Supervised Inference} \\
% \end{tabular}
% \vspace{-0.1cm}
% \caption{Structures of our proposed sequential probabilistic models. Sequence $\vx_{1:T}$ is disentangled into static part $\vz^c$ and dynamic parts $\{\vz^m_t\}$. %$\vh_t$ in (a, c) is hidden state of an LSTM and $\vh_t$ in (b, d) is extracted feature from encoder network; 
% (a) sequence is generated by randomly sampling $\{\vz^c,\vz_t^m\}$ from priors and concatenating them as input into an LSTM to get hidden state $\vh_t$ for the decoder; (b) $\vz^c$ is inferred from $\vx_{1:T}$, and $\vz^m_t$ is inferred from $\vh_t$ and $\vz^{m}_{t-1}.$; (c) is the same as (a) except concatenating additional categorical $\va.$; (d) A categorical latent variable $\va$ is inferred from the dynamic latent codes. The detailed structures of the encoder and decoder are in the supplementary material.\label{fig:struct}}
% \end{figure*}
\begin{figure*}[ht]
\centering
\resizebox{1\textwidth}{!}{
\begin{tabular}{cccc}\hspace{-0.2cm}
\includegraphics[height=0.166\textwidth]{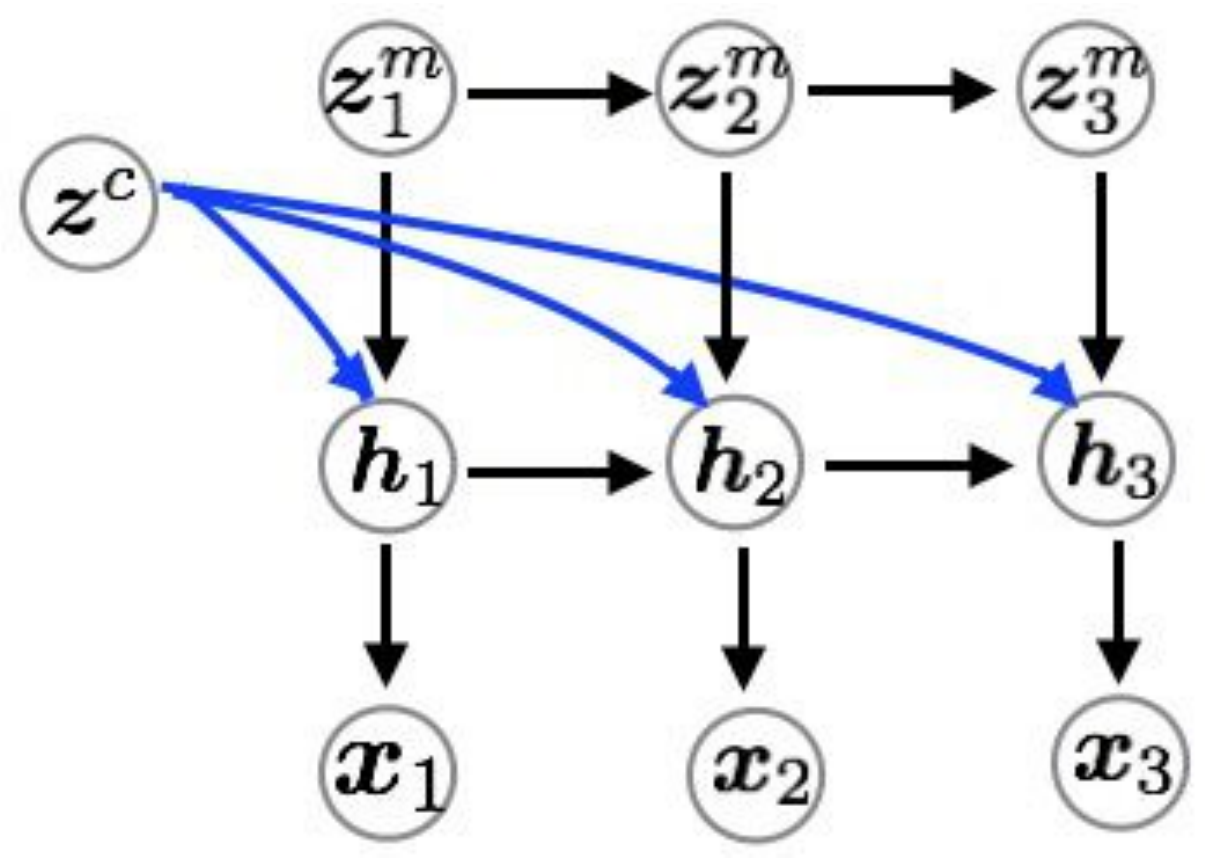}& \hspace{-0.55cm}
\includegraphics[height=0.166\textwidth]{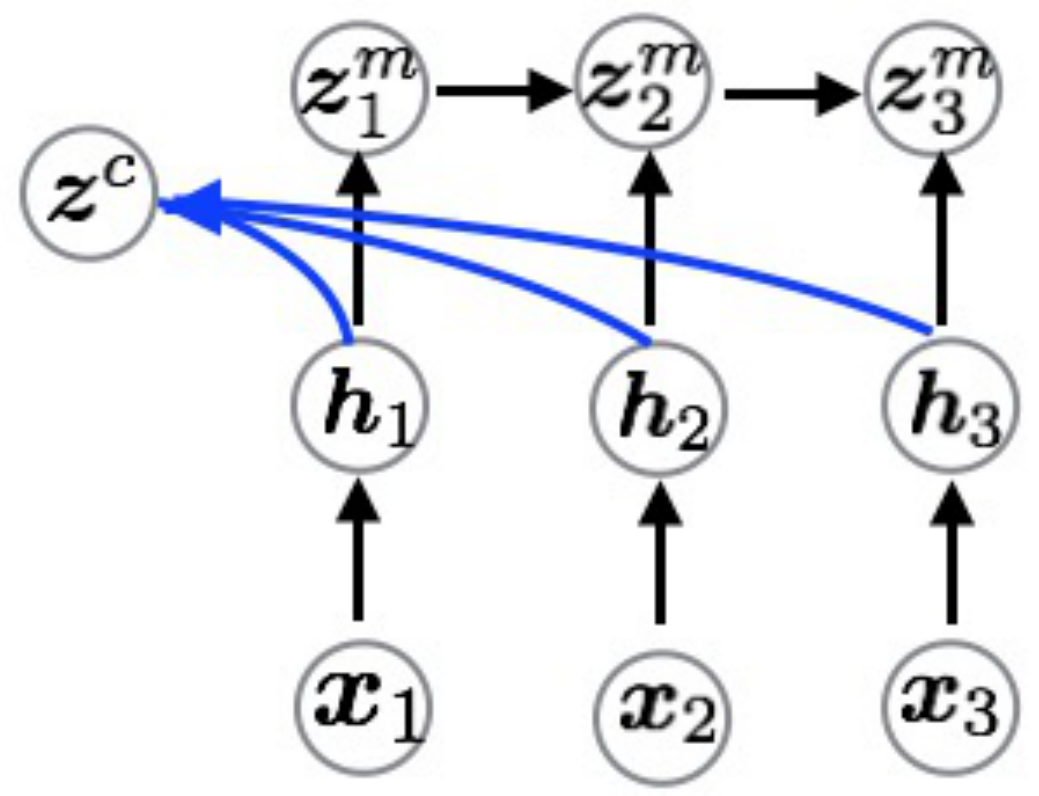}& \hspace{-0.55cm}
\includegraphics[height=0.166\textwidth]{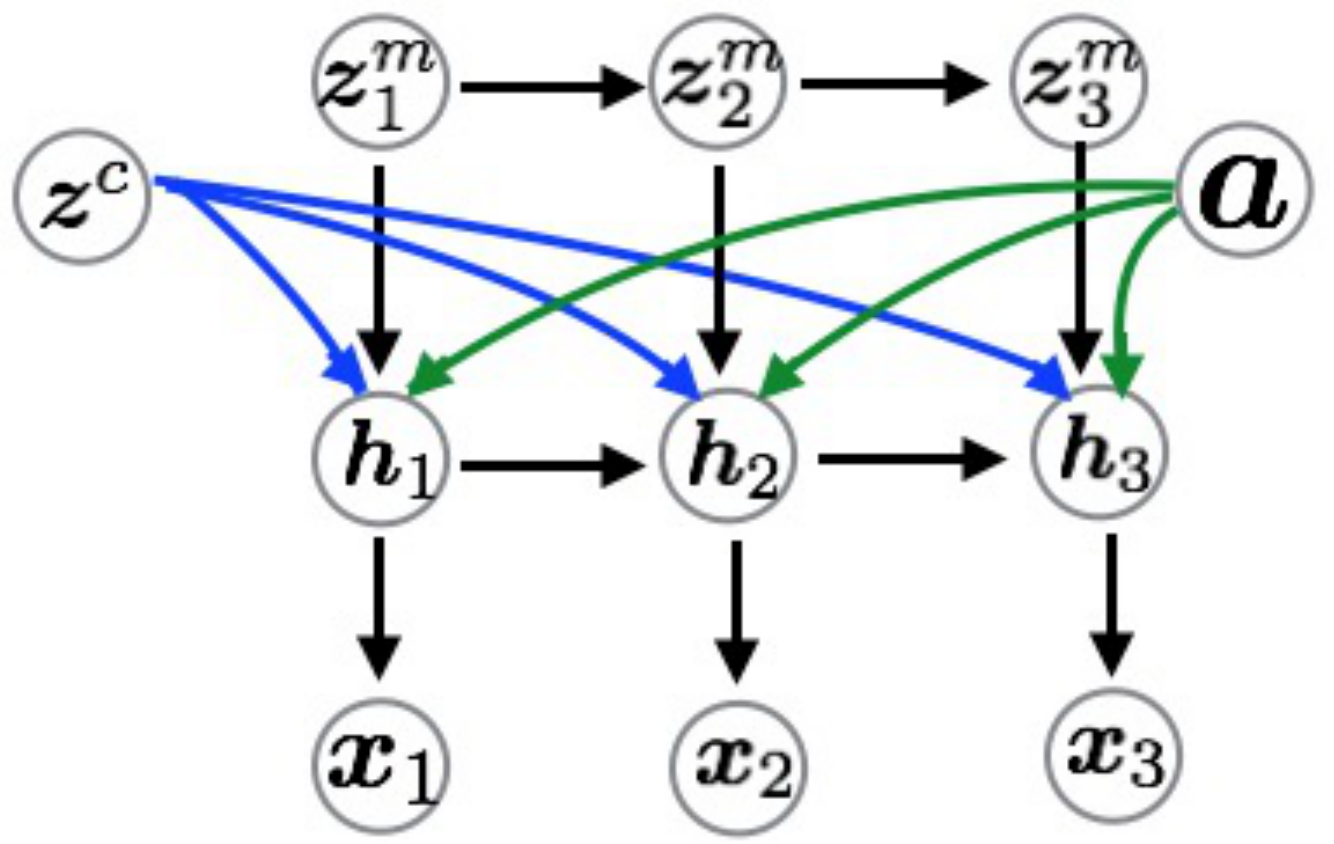}& \hspace{-0.55cm}
\includegraphics[height=0.166\textwidth]{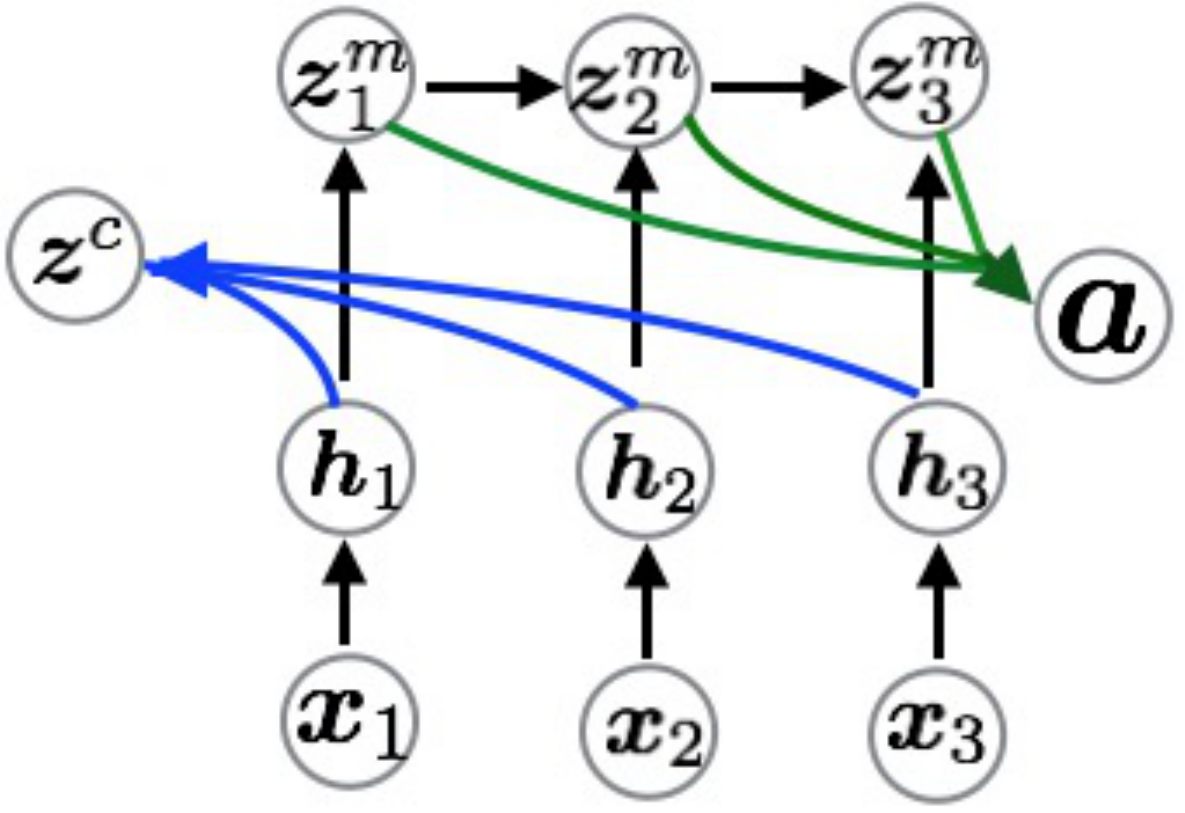}\\
\vspace{-0.2cm}
{\small (a) Generative Model} & {\small\! (b) Inference Model} &  {\small \!\! \! \!\! (c) Weakly Supervised Generative}& {\small \!\!\!\!\! (d) Weakly Supervised Inference} \\
\end{tabular}}
\vspace{-0.1cm}
\caption{Structures of our proposed sequential probabilistic models. Sequence $\vx_{1:T}$ is disentangled into static part $\vz^c$ and dynamic parts $\{\vz^m_t\}$. %$\vh_t$ in (a, c) is hidden state of an LSTM and $\vh_t$ in (b, d) is extracted feature from encoder network; 
(a) Sequence is generated by randomly sampling $\{\vz^c,\vz_t^m\}$ from priors and concatenating them as input into an LSTM to get hidden state $\vh_t$ for the decoder; (b) $\vz^c$ is inferred from $\vx_{1:T}$ with an LSTM, and $\vz^m_t$ is inferred from $\vh_t$ and $\vz^{m}_{t-1}$ with another LSTM; (c) is the same as (a) except concatenating additional categorical $\va$; (d) A categorical latent variable $\va$ is inferred from the dynamic latent codes. The detailed structures of the encoder and decoder are in the supplementary material.\label{fig:struct}}
\end{figure*}

Given a high-dimensional sequence $\vx_{1:T}$, our goal is to learn a disentangled representation of time-invariant latent code $\vz^c$ and time-variant latent code $\vz^m_t$, along the sequence. Let $\vz_t=(\vz^c, \vz^m_t)$ be the latent code of $\vx_t$. Let $X_t$, $Z_t$, $Z^c$ and $Z^m_t$ be random variables with realizations $\vx_t$, $\vz_t$, $\vz^c$ and $\vz^m_t$ respectively, and denote $\CD=X_{1:T}.$ To achieve this goal, we define the following probabilistic generative model by assuming $Z^m_t$ and $Z^c$ are independent,
\begin{equation}
\label{def:gen}
P(X_{1:T}, Z_{1:T})= P(Z^c)\prod_{t=1}^T P_\psi(Z^m_t|Z^m_{<t})P_{\theta}(X_t|Z_t), 
\end{equation}
where $P(Z_{1:T})=P(Z^c)\prod_{t=1}^T P_\psi(Z^m_t|Z^m_{<t})$ is the prior in which $Z_t=(Z^c,Z^m_t)$, and the decoder model $P_{\theta}(X_t \mid Z_t)$ is a Dirac delta distribution. In practice, $P(Z^c)$ is chosen as $\mathcal{N}(\bd{0}, \bd{I})$ and $P_\psi(Z^m_t|Z^m_{<t})=\mathcal{N}(\bd{\mu}_{\psi}(Z^m_{<t}), \bd{\sigma}_{\psi}^2(Z^m_{<t}))$, $\bd{\mu}_{\psi}$ and $\bd{\sigma}_{\psi}$ are  parameterized by Recurrent Neural Networks (RNNs). 
The inference model $Q$ is defined as 
\vspace{-0.2cm}
\begin{equation}
\label{def:inf}
Q_\ff(Z^c, Z_{1:T}^m|X_{1:T}) = Q_\ff(Z^c|X_{1:T}) \prod_{t=1}^T Q_\ff(Z_{t}^m\mid Z_{<t}^m, X_t),   
\end{equation}
where $Q_\ff(Z^c|X_{1:T})$ and $Q_\ff(Z_{t}^m\mid Z_{<t}^m, X_t)$ are also Gaussian distributions parameterized by RNNs. The structures of the generative model~\eqref{def:gen} and the inference model~\eqref{def:inf} are provided in Fig.~\ref{fig:struct}. 

% We use a convolutional LSTM for the generator $\vx_t=G(\vx_{1:t-1},\mathbf{z}_{1:t})$ and $q(\mathbf{z}_t|\mathbf{x}_{1:t})$. Frames are encoded via a feed-forward convolutional network and then are fed to the LSTM model. 
% \begin{equation}
% h_t=Enc(\mathbf{x}_t)
% g_t=LSTM_\theta(h_{t-1},\mathbf{z}_t)
% \end{equation}

% $$\max_{\gamma} \E_{X}[\log D_{\gamma}(\vz^c)+\log(1-D_{\gamma}(\tilde{\vz^c}))]$$
% Update $p_{\theta}$, prior $p_{\psi}$, $q_{\phi}$ by descending:
% $$\sum_{t=1}^T \E_X[c(\vx_t, G(\tilde{\vz}^c, \tilde{\vz}_t^m)) + \mathrm{MMD}(\vz_t^m, \tilde{\vz}_t^m)].$$

% \begin{equation}
% \mathrm{D}_{\mathrm{WVAE}}(P_X, P_G):= \sum_{t=1}^T \inf_{q_{\phi}(\mathbf{z}_t|\mathbf{x}_{1:t})} \mathbb{E}_{P_X} \mathbb{E}_{q_{\phi}(z_t|x_{1:t})}[p_\vthe(\vx_t\mid\vz_t)]- \KL(q_\ff(\vz_t|\vx_{1:t})\mid\mid p_\psi(\vz_t\mid\vx_{1:t-1}))) + \beta \mathbb{D}(Q_Z, P_Z),  \end{equation}
% where $Q_Z$ is the marginal distribution of Z and $P_Z$ is prior distribution.

\subsection{R-WAE minimizes a penalized form of a Wasserstein distance} The optimal transport cost between two distributions $P_\CD$ and $P_G$ with respective sequential variables $X_{1:T}$ ($X_{1:T}\!\!  \sim\!\!   P_\CD$) and $Y_{1:T}$ ($\! Y_{1:T}\!\!  \sim\!\!   P_G$) is given as follows,
\begin{equation}
W(P_\CD, P_G) := \inf_{\Gamma \sim \cp(X_{1:T}\sim P_\CD, Y_{1:T}\sim P_G)}\E_{(X_{1:T}, Y_{1:T})\sim \Gamma}[c(X_{1:T}, Y_{1:T})],   
\end{equation}
$\cp(X_{1:T}\!\!  \sim\!\!   P_\CD, \! Y_{1:T}\!\!  \sim\!\!   P_G)$ is a set of all joint distributions with marginals $P_\CD$ and $P_G$ respectively.

When we choose $c(\vx,\vy)=\|\vx-\vy\|^2$ (2-Wasserstein distance) and $c(X_{1:T}, Y_{1:T})=\sum_t \|X_t-Y_t\|^2$ by linearity, it is easy to derive the optimal transport cost for disentangled sequential variables.
%erran: fixed. 
%\textcolor{red}{ACs think we should label this as Theorem 1}
%\begin{thm}
\begin{theorem}
With deterministic $P(X_t|Z_t)$ and any function $Y_t=G(Z_t),$ we derive
\begin{equation}
W(P_\CD, P_G) =\inf_{Q: Q_{Z^c}= P_{Z^c}, Q_{Z_{1:T}^m }=P_{Z_{1:T}^m }}\sum_t \E_{P_\CD}\E_{Q(Z_t|Z_{<t}, X_t)}[c(X_t, G(Z_t))],     
\end{equation}
where $Q_{Z_{1:T}}=Q_{Z^c}Q_{Z_{1:T}^m}$ is the marginal distribution of $Z_{1:T}$ when $X_{1:T}\sim P_\CD$ and $Z_{1:T}\sim Q(Z_{1:T}|X_{1:T})$ and $P_{Z_{1:T}}$ is the prior.
Based on the assumptions, we have an upper bound, 
\begin{equation}
W(P_\CD, P_G) \le \inf_{Q\in \mathcal{S}} \sum_t \E_{P_\CD}\E_{Q(Z_t|Z_{<t}, X_t)}[c(X_t, G(Z_t))],    
\end{equation}
where the subset $\mathcal{S}$ is $\mathcal{S}=\{Q: Q_{Z^c}= P_{Z^c}, Q_{Z_{1}^m}=P_{Z_{1}^m}, Q_{Z_{t}^m|Z_{<t}^m} = P_{Z_{t}^m|Z_{<t}^m}\}$ .
\label{thm:opt}
%\end{thm}
\end{theorem}

In practice, we have the following objective function of our proposed R-WAE based on Theorem~\ref{thm:opt},
\begin{equation}
\label{obj:loss}
\sum_{t=1}^T \E_{Q(Z_t|Z_{<t}, X_t)}[c(X_t, G(Z_t))]+\beta_1~ \bbd(Q_{Z^c}, P_{Z^c}) + \beta_2 \sum_{t=1}^T  \bbd (Q_{Z^m_t|Z^m_{<t}}, P_{Z^m_t|Z^m_{<t}}), 
\end{equation}
where $\bbd$ is an divergence between two distributions, and the second and third terms are, respectively, regularization terms for $Z^c$ and $Z^m_t$. In the following, we will present two different approaches to calculating the regularization terms in section 3.2 and 3.3. Because we cannot straightforwardly estimate the marginals $Q_\ff (Z^c)$ and $Q_{\ff}(Z_t^m|Z_{<t}^m),$ we cannot directly use KL divergence in the two regularization terms, but we can optimize the RHS of \eqref{obj:loss} by likelihood-free optimizations \citep{gretton2007kernel, goodfellow2014generative, nowozin2016f, arjovsky2017wasserstein} when samples from all distributions are available.

\subsection{$\bbd_{\JS}$ Penalty for $Z^c$ and $\MMD$ Penalty for $Z^m$} The prior distribution of $Z^c$ is chosen as a multivariate unit-variance Gaussian, $\mathcal{N}(\bd{0}, \bd{I})$. We can choose penalty $\bbd_{\JS}(Q_{Z^c},\! P_{Z^c})$ for $Z^c$
and apply min-max optimization by introducing a discriminator $D_{\gamma}$~ \citep{goodfellow2014generative}. Instead of performing optimizations in high-dimensional input data space, we move the adversarial optimization to the latent representation space of the content with a much lower dimension. Because the prior distribution of $\{Z_t^m \}$ is dynamically learned during training, it is challenging to use $\bbd_{\JS}$ to regularize $\{Z^m_t\}$ with a discriminator, which will induce a third minimization within a min-max optimization. Therefore, we use $\MMD$ to regularize $\{Z^m_t\}$ as samples from both distributions are easy to obtain (dimension of $\vz^m_t$ is less than $20$ in our experiments on videos). With a kernel $k$, $\MMD_k(Q, P)$ is approximated by samples from $Q$ and $P$~\citep{gretton2007kernel}. The regularization terms can be summarized as follows and we call the resulting model R-WAE(GAN) (see Algorithm \ref{alg:alg1} in Appendix for details): 
\begin{equation}
\label{pen:gan}
\bbd(Q_{Z^c},\! P_{Z^c})=\bbd_{\JS}(Q_{Z^c},\! P_{Z^c});~\bbd (Q_{Z^m_t|Z^m_{<t}},\! P_{Z^m_t\!|\!Z^m_{<t}})= \MMD_{k}(Q_{Z^m_t|Z^m_{<t}},\! P_{Z^m_t\!|\!Z^m_{<t}}).      
\end{equation}

\subsection{Scaled $\MMD$ Penalty for $Z^c$  and $\MMD$ Penalty for $Z^m$} $\MMD$ with neural kernels for generative modeling of real-world data \citep{li2017mmd, binkowski2018demystifying, arbel2018gradient} motivates us to use only $\MMD$ as regularization in Eq.~\eqref{obj:loss},  
\begin{equation}
\label{pen:mmd}
\bbd(Q_{Z^c},\! P_{Z^c})=\MMD_{k_{\gamma}} (Q_{Z^c},\! P_{Z^c}); ~\bbd (Q_{Z^m_t|Z^m_{<t}},\! P_{Z^m_t\!|\!Z^m_{<t}})= \MMD_{k}(Q_{Z^m_t|Z^m_{<t}},\! P_{Z^m_t\!|\!Z^m_{<t}}),    
\end{equation}
where $k_\gamma$ is a parametrized family of kernels \citep{li2017mmd, binkowski2018demystifying, arbel2018gradient} defined as $k_\gamma(\vx, \vy)=k(f_\gamma(\vx), f_\gamma(\vy))$ and $f_\gamma(\vx)$ is a feature map, which is more expressive and used for $Z^c$ with equal or higher dimension than $Z_t^m$. The details of optimizing the first term $\MMD_{k_\gamma} (Q_{Z^c},\! P_{Z^c})$ in Eq.~\eqref{pen:mmd} is provided in Appendix D based on scaled MMD~\citep{arbel2018gradient}, a principled and stable technique for training $\MMD$-based critic. We call the resulting model R-WAE(MMD) (see Algorithm~\ref{alg:alg2} in Appendix for details).

\subsection{Weakly Supervised Disentanglement with a Known Number of Actions} When the number of actions (motions) in sequential data, denoted by $A$, is available, we incorporate a categorical latent variable $\va$ (a one-hot vector whose dimension is $A$) to enhance the disentanglement of the dynamic latent codes of the motions. The inference model for $\va$ is designed as $q_{\phi}(\va|\vx_{1:T}, \vz_{1:T}^m).$ Intuitively, the action is inferred from the motion sequence to recognize its label. Learning such a  categorical distribution requires a continuous relaxation of the discrete random variable in order to backpropagate its gradient. Let $\alpha_1,\cdots,\alpha_A$ be the class probabilities, we can obtain a sample $\wt{\va}=(y_1,\cdots, y_A)$ from its continuous relaxation by first sampling $\bd{g}=(g_1,\cdots, g_A)$ with $g_j\sim\mathrm{Gumbel}(0,1)$ and then applying transformation $\wt{a}_j = \exp((\log \alpha_j+g_j)/\tau)\sum_i\exp((\log \alpha_i+g_i)/\tau),$
where $\tau$ is a temperature parameter controlling the approximation. To learn the categorical distribution using the reparameterization trick, we use a regularizer $\bbd_{\KL}(q_\phi(\wt{\va}|\vx_{1:T}, \vz_{1:T}^m), p(\wt{\va}))$, where $p(\wt{\va})$ is a uniform Gumbel-Softmax prior distribution~\citep{jang2016categorical, maddison2016concrete}. The motion variable is augmented as $\vz^R_t=(\vz^m_t, \va)$, and learning joint continuous and discrete latent representation of image data has been extensively discussed in~\citep{dupont2018learning} (see Fig.~\ref{fig:struct}(c,d) for illustrations).

\section{Analyzing R-WAE from an Information Theory Perspective}
%erran: fixed. \textcolor{red}{ACs think we should label this as Theorem 2}
\begin{theorem}
%\begin{thm}
\label{thm:milower}
If the mutual information (MI) between $Z_{1:T}$ and $X_{1:T}$ is defined in terms of the inference model $Q$, $I(Z_{1:T}; X_{1:T})  = \E_{Q(X_{1:T}, Z_{1:T})}[\log Q(Z_{1:T}| X_{1:T})-\log Q(Z_{1:T})],$ where $Q(X_{1:T}, Z_{1:T})=Q(Z_{1:T} |X_{1:T})P(X_{1:T})$ and $Q(Z_{1:T})=\sum_{X_{1:T}} Q(X_{1:T}, Z_{1:T})$, we have a lower bound:
\vspace{-0.2cm}
\begin{align}
\label{mi:lower}
 I(Z_{1:T}; X_{1:T})& 
 \! \ge\!\! \sum_{t=1}^T  \E_{P_\CD}\!\big [\E_{Q_\ff}[\log P_{\theta}(X_t\!\mid\!\! Z_t)\!\!-\!\!\log P(\CD)]\! -\! \E_{Q_\ff (Z^c|\! X_{1:T})}[ \log Q_{\ff}(Z^c)\! -\!\!\log P(Z^c)]\big ] \notag \\
 &\!-\!\!\sum_{t=1}^T \E_{P_\CD}\big [\E_{Q_\ff(Z_t^m|Z_{<t}^m, X_t)}[ \log Q_{\ff}(Z_t^m|Z_{<t}^m) \!\!-\!\! \log P(Z_t^m|Z_{<t}^m)\!\big].
\end{align}
\vspace{-0.2cm}
%\end{thm}
\end{theorem}
Theorem~\ref{thm:milower} shows that R-WAE maximizes a lower bound of the mutual information between $X_{1:T}$ and $Z_{1:T}$, which theoretically guarantees that R-WAE learns semantically meaningful latent representations of input sequences. With constant removed, the RHS of \eqref{obj:loss} and \eqref{mi:lower} are the same if $\bbd$ is KL divergence. In spite of being theoretically important, Theorem~\ref{thm:milower} with KL divergence cannot be directly used for the regularization terms of R-WAE in practice, because we cannot straightforwardly estimate the marginals $Q_\ff (Z^c)$ and $Q_{\ff}(Z_t^m|Z_{<t}^m)$ as discussed previously. 

From Eq.~\eqref{obj:loss} and \eqref{mi:lower}, we can obtain the following theorem. %\textcolor{red}{Hi Jun, can you please put the proof in the Appendix?}
%erran: fixed. \textcolor{red}{ACs think we should label this as Theorem 3}
\begin{theorem}
%\begin{thm}
\label{lem:reg}
 When its distribution divergence is chosen as $\KL$ divergence, the regularization terms in Eq.~\eqref{obj:loss} jointly minimize the $\KL$ divergence between the inference model $Q(Z_{1:T}|X_{1:T})$ and the prior model $P(Z_{1:T})$ and maximize the mutual information between $X_{1:T}$ and $Z_{1:T},$
% \begin{align}
% \label{eq:reg}
% \KL(Q(Z^c_{1:T})||P(Z^c_{1:T})) & = E_{p_{\CD}}[\KL(Q(Z_{1:T}^c|X_{1:T})||P(Z^c_{1:T}))] - I(X_{1:T}, Z^c_{1:T}),  \\ \notag
% \KL(Q(Z^m_{1:T})||P(Z^m_{1:T})) & = 
% E_{p_{\CD}}[\KL(Q(Z_{1:T}^m|X_{1:T})||P(Z^m_{1:T}))] - I(X_{1:T}, Z^m_{1:T}) .    
% \end{align}
\vspace{-.5cm}

\begin{align}
\label{eq:reg}
\KL(Q(Z^c)||P(Z^c)) & = \E_{p_{\CD}}[\KL(Q(Z^c|X_{1:T})||P(Z^c))] - I(X_{1:T};Z^c),  \\
\!\! \KL(Q(Z_{t}^m|Z_{<t}^m)||P(Z_{t}^m|Z_{<t}^m)) & \!  = \! 
\E_{p_{\CD}}[\KL(Q(Z_{t}^m|Z_{<t}^m, X_{1:T})||P(Z_{t}^m|Z_{<t}^m)] - I(X_{1:T}; Z_{t}^m|Z_{<t}^m),   \nonumber  
\end{align}
where the mutual information is defined in terms of the inference model as in Theorem \ref{thm:milower}.
%\end{thm}
\end{theorem}
Theorem \ref{lem:reg} shows that, even if adopting KL divergence, the regularization in the loss of R-WAE still improves over the one in vanilla VAE, which only has the first term in the RHS of Eq.~\eqref{eq:reg}. 
%\textcolor{red}{Hi Erran and Jun, please write down what we discussed here why R-WAE is much better than DS-VAE due to the two MI terms}
The two mutual information terms explicitly enforce mutual information maximization between input data and unexchangeable disentangled latent representations, $Z^c$ and $Z_{t}^m$. Therefore, R-WAE is superior to recurrent VAE (DS-VAE).

\vspace{-.4cm}
\begin{figure*}
\begin{center}
\begin{subfigure}{0.33333\textwidth}
\centering
\begingroup
\begin{tabular}{cccc}
\includegraphics[height=.185\textwidth]{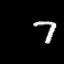} & \hspace{-0.4cm}
\includegraphics[height=.185\textwidth]{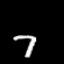}& \hspace{-0.4cm}
\includegraphics[height=.185\textwidth]{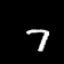}& \hspace{-0.4cm}
\includegraphics[height=.185\textwidth]{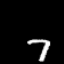} \\
\includegraphics[height=.185\textwidth]{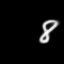}& \hspace{-0.4cm}
\includegraphics[height=.185\textwidth]{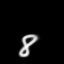}& \hspace{-0.4cm}
\includegraphics[height=.185\textwidth]{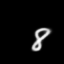}& \hspace{-0.4cm}
\includegraphics[height=.185\textwidth]{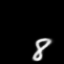} \\
\includegraphics[height=.185\textwidth]{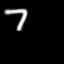}& \hspace{-0.4cm}
\includegraphics[height=.185\textwidth]{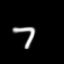}& \hspace{-0.4cm}
\includegraphics[height=.185\textwidth]{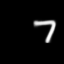}& \hspace{-0.4cm}
\includegraphics[height=.185\textwidth]{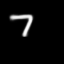} \\
\includegraphics[height=.185\textwidth]{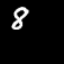}& \hspace{-0.4cm}
\includegraphics[height=.185\textwidth]{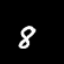}& \hspace{-0.4cm}
\includegraphics[height=.185\textwidth]{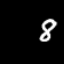}& \hspace{-0.4cm}
\includegraphics[height=.185\textwidth]{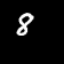} \\
\end{tabular}
\endgroup
\vspace{-0.2cm}
\caption{t=1;~~~ t=10;~~~ t=50;~~~ t=100} 
\end{subfigure}
\hspace{-.3cm}
\begin{subfigure}{0.33333\textwidth}
\centering
\begingroup
\begin{tabular}{cccc}
\includegraphics[height=.185\textwidth]{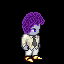}& \hspace{-0.4cm}
\includegraphics[height=.185\textwidth]{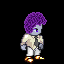}& \hspace{-0.4cm}
\includegraphics[height=.185\textwidth]{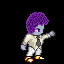}& \hspace{-0.4cm}
\includegraphics[height=.185\textwidth]{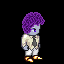}\\
\includegraphics[height=.185\textwidth]{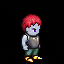}&\hspace{-0.4cm}
\includegraphics[height=.185\textwidth]{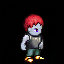}&\hspace{-0.4cm}
\includegraphics[height=.185\textwidth]{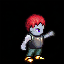}&\hspace{-0.4cm}
\includegraphics[height=.185\textwidth]{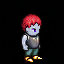}\\ 
\includegraphics[height=.185\textwidth]{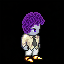}& \hspace{-0.4cm}
\includegraphics[height=.185\textwidth]{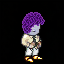}& \hspace{-0.4cm}
\includegraphics[height=.185\textwidth]{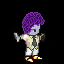}& \hspace{-0.4cm}
\includegraphics[height=.185\textwidth]{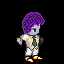}\\ 
\includegraphics[height=.185\textwidth]{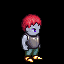}& \hspace{-0.4cm}
\includegraphics[height=.185\textwidth]{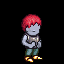}& \hspace{-0.4cm}
\includegraphics[height=.185\textwidth]{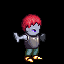}& \hspace{-0.4cm}
\includegraphics[height=.185\textwidth]{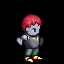}\\ 
\end{tabular}
\endgroup
\vspace{-0.2cm}
\caption{t=1;~~~~~ t=3;~~~~~ t=5;~~~~~ t=7}
\end{subfigure}
\hspace{-.3cm}
\begin{subfigure}{0.33333\textwidth}
\centering
\begingroup
\begin{tabular}{cccc}
 \includegraphics[height=.185\textwidth]{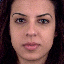}& \hspace{-0.4cm}
\includegraphics[height=.185\textwidth]{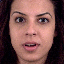}& \hspace{-0.4cm}
\includegraphics[height=.185\textwidth]{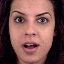}& \hspace{-0.4cm}
\includegraphics[height=.185\textwidth]{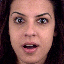}\\
\includegraphics[height=.185\textwidth]{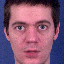}& \hspace{-0.4cm}
\includegraphics[height=.185\textwidth]{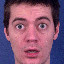}& \hspace{-0.4cm}
\includegraphics[height=.185\textwidth]{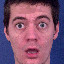}& \hspace{-0.4cm}
\includegraphics[height=.185\textwidth]{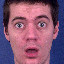} \\
\includegraphics[height=.185\textwidth]{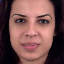}&\hspace{-0.4cm}
\includegraphics[height=.185\textwidth]{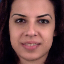}&\hspace{-0.4cm}
\includegraphics[height=.185\textwidth]{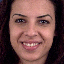}&\hspace{-0.4cm}
\includegraphics[height=.185\textwidth]{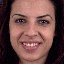} \\
\includegraphics[height=.185\textwidth]{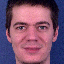}& \hspace{-0.3cm}
\includegraphics[height=.185\textwidth]{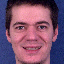}& \hspace{-0.3cm}
\includegraphics[height=.185\textwidth]{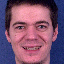}& \hspace{-0.3cm}
\includegraphics[height=.185\textwidth]{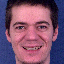} \\
\end{tabular}
\endgroup
\vspace{-0.2cm}
\caption{t=1;~~~~~ t=3;~~~~~ t=5; ~~~~~t=7} 
\end{subfigure}
\vspace{-0.2cm}
\caption{Illustration of disentangling the motions and contents of two videos on the test data of SM-MNIST ($T=100$), Sprites ($T=8$) and MUG dataset ($T=8$). The first row and fourth row are original videos. The second row and third row are generated sequences by swapping the respective motion variables while keeping content variable the same (sampled at 4 time steps for illustrations). \label{fig:distoy}}
\end{center}
\end{figure*}

\section{Experiments}
We conduct extensive experiments on four datasets to quantitatively and qualitatively validate our methods.  %We compare our methods with 
The baseline methods for comparisons are DS-VAE~\citep{li2018disentangled} and MoCoGAN~\citep{tulyakov2018mocogan}. We train our models on Stochastic moving MNIST (SM-MNIST), Sprites, and TIMIT datasets under a completely unsupervised setting. The number of actions (motions) is used as prior information for all methods on MUG facial dataset. The detailed descriptions of datasets, architectures, and hyperparameters are provided in Appendix C, D, and G, respectively.  

\subsection{Qualitative Results on Disentanglement}
%\paragraph{Qualitative Results on SM-MNIST and Sprites Datasets} 
We encode two original videos on the first and fourth row in Fig.~\ref{fig:distoy} and generate videos on the second and third row by swapping corresponding $\{\vz^c\}$ and $\{\vz^m_{1:T}\}$ between videos for style transfer. Fig.~\ref{fig:distoy}(left) shows that even testing on the long sequence (trained with $T=100$), our R-WAE can disentangle content and motions exactly. In Fig.~\ref{fig:distoy}(right), we do the same swapping on Sprites. We can see that the generated swapped videos have exact same appearances and actions as the corresponding original ones. On the MUG dataset, it is interesting to see that we can swap different motions between different persons.     
\begin{minipage}{\textwidth}
\begin{minipage}[h]{0.41\textwidth}
\begin{center}
\begin{tabular}{|p{1.68cm}|p{1.39cm}|p{1.43cm}|}
\hline
 \multirow{2}{*}{\!\!\!\backslashbox{\! {\small Methods}}{\!\!\!\!\!\!\!\!\!\!\!\!\!\!\!\!\! {\small \!\! Datasets}}}
& \multicolumn{2}{|c|}{~~EER~} \\ 
 \cline{2-3}
 & \!\! \small{$\vz^c=16$} \!\! $\downarrow$ &\!\!\small{$\vz^m=16$} $\uparrow$\\ \hline
{\small FHVAE} &  5.06\% & 22.77\%  \\ \hline
{\small DS-VAE} &  5.64\% & 19.20\% \\ \hline
{\small  R-WAE} & {\bf 4.73\%} & {\bf 23.41\%} \\ \hline
\end{tabular}
\captionof{table}{EER on TIMIT speech dataset under the same dimension setting of segment-level $\vz^c$ and sequence-leve $\vz^m$ for FHVAE~\citep{hsu2017unsupervised}, DS-VAE (full q)~\citep{li2018disentangled} and R-WAE(MMD), respectively. Small EER is better for $\vz^c$ and larger EER is better for $\vz^m$. \label{table:speech}}
\end{center}
\end{minipage}
\hspace{0.1cm}
\begin{minipage}[bth]{0.56 \textwidth}
    \centering
  \begin{tabular}{|p{1.65cm}|p{.8cm}|p{1.32cm}|p{.6cm}|p{.6cm}|p{.4cm}|}
 \hline
  \multirow{2}{*}{\!\!\!\backslashbox{\! {\small Methods}}{\!\!\!\!\!\!\!\!\!\!\!\!\!\!\!\!\! {\small \!\! Datasets}}}& \multicolumn{2}{|c|}{ Sprites}& \multicolumn{1}{|c|}{{\small SM-MNIST}} \\
 \cline{2-4}
 & actions & {\small content} & digits \\
 \hline
{\small DS-VAE(S)} &  8.11\%  & 3.98\%  & 3.31\% \\ \hline
R-WAE(S) & {\bf 5.83\%} & {\bf 2.45\%} & {\bf 1.78\%} \\
 \hline
{\small DS-VAE(C)} & 10.37\%  & 4.86\% & 4.26\% \\ \hline
R-WAE(C) & {\bf 7.72\%} & {\bf 3.31\%} & {\bf 2.84\%} \\
 \hline
\end{tabular}  
\captionof{table}{Comparison of averaged classification errors. On Sprites dataset, fix one encoded attribute and randomly sample others. On SM-MNIST dataset, we fix the encoded $\vz^c$ and randomly sample the motion sequence from the learned prior $p_{\psi}(\vz_t^m|\vz_{<t}^m).$ We cannot quantitatively verify the motion disentanglement on SM-MNIST.\label{toy:quant}}
\end{minipage}
\end{minipage}

\vspace{-.6cm}
\subsection{Quantitative Results}
\vspace{-.3cm}
\paragraph{SM-MNIST and Sprites Datasets} We quantitatively evaluate the disentanglement of our R-WAE(MMD). In Table ~\ref{toy:quant}, "S" denotes a simple encoder/decoder architecture, where the encoders in both our model and DS-VAE~\citep{li2018disentangled} only use 5 layers of convolutional and deconvolutional networks adopted from DS-VAE~\citep{li2018disentangled}. "C" denotes a complex encoder/decoder architecture where we use Ladder network~\citep{sonderby2016ladder, zhao2017learning} and ResBlock~\citep{he2016deep}, provided in Appendix E. On SM-MNIST, we get the labeled latent codes $\{\vz^c\}$ of test videos $\{\vx_{1:T}\}$ with $T=10$ and randomly sample motion variables $\{\vz^m_{1:T}\}$ to get labeled new samples. We pretrain a classifier and predict the accuracy on these labeled new samples. The accuracy on SM-MNIST dataset is evaluated on 10000 test samples. On Sprites, the labels of each attribute(skin colors, pants, hair styles, tops and pants) are available. We get the latent codes by fixing one attribute and randomly sample other attributes. We train a classifier for each attribute and evaluate the disentanglement of each attribute. The accuracy is based on $296\times 9$ test data. Both DS-VAE and R-WAE(MMD) have extremely high accuracy (99.94\%) when fixing hair style attribute, which is not provided in Table~\ref{toy:quant} due to space limit. As R-WAE(GAN) and R-WAE(MMD) have similar performance on these datasets, we only provide the results and parameters of R-WAE(MMD) to save space. There are two interesting observations in Table ~\ref{toy:quant}. First, the simple architecture has better disentanglement than the complex architecture overall. The reason is that the simple architecture has sufficient ability to extract features and generate clear samples to be recognized by the pretrained classifiers. But the simple architecture cannot generate high-quality samples when applied to real data. Second, our proposed R-WAE(MMD) achieve better disentanglement than DS-VAE~\citep{li2018disentangled} on both corresponding architectures. The attributes within content latent variables are independent, our model can further disentangle the factors. Compared to DS-VAE, these results demonstrate the advantages of R-WAE with implicit mutual information maximization terms. Due to space limit, we also include similar comparisons on a new Moving-Shape dataset in Appendix I. As the number of possible motions in SM-MNIST is infinite and random, we cannot evaluate the disentanglement of motions by training a classifier. We fix the encoded motions $\{\vz^m_{1:T}\}$ and randomly sample content variables $\{\vz^c\}.$ We also randomly sample a motion sequence $\{\vz^m_{1:T}\}$ and randomly sample contents $\{\vz^c\}$. We manually check the motions of these samples and almost all have the same corresponding motion even though the sequence is long ($T=100$).  
\vspace{-.4cm}
\paragraph{TIMIT Speech Dataset} We quantitatively compare our R-WAE with FHVAE and DS-VAE on the speaker verification task under the same setting as \citep{hsu2017unsupervised,li2018disentangled}. The evaluation metric is based on equal error rate (EER), which is explained in detail in Appendix C. The lower EER on  $\vz^c$ encoding the timbre of speakers is better and the higher EER on $\vz^m$ encoding linguistic content is better. From Table~\ref{table:speech}, our model can disentangle $\vz^c$ and  $\vz^m$ well. We can see that our R-WAE(MMD) has the best EER performance on both content attribute and motion attribute. In Appendix H we show by style transfer experiments that the learned dynamic factor encodes semantic content which is comparable to DS-VAE.

% Ligong: following is the original table with backslashbox
\begin{table*}
\begin{center}
\resizebox{1\textwidth}{!}{
  \begin{tabular}{|p{2.2cm}|p{1.5cm}|p{2.2cm}|p{1.9cm}|p{2.22cm}|p{2.2cm}|} \hline
\backslashbox{{\small Metrics}}{\!\!\!\!\!\!\!\!\!\!\!\!\!\!\!\!\!\! {\small Methods}} &{\small MocoGAN} & {\small DS-VAE(NA)}& {\small DS-VAE(W)} &{\small R-WAE(MMD)}&{\small R-WAE(GAN)}\\ \hline
 {\small Accuracy ~$\uparrow$} & 75.50\%  & 66.73\% & 82.84 \% & 88.62\% & {\bf 90.15\%}   \\ \hline
 {\small Intra-entropy$\downarrow$} & 0.26 & 0.28& 0.23 & 0.17 & {\bf 0.15} \\ \hline
{\small Inter-entropy$\uparrow$} & 1.78 & 1.77 & 1.78 & {\bf 1.79} & {\bf 1.79}\\ \hline
{\small Inception Score~$\uparrow$} & 4.60 & 4.44 & 4.71 & 5.05 & {\bf 5.16}  \\ \hline
{\small ~~~ FID~$\downarrow$} & 16.95& 18.72&14.79& {\bf 12.21}& {\bf 10.86} \\
\hline
 \end{tabular}}
\vspace{-0.3cm}
 \caption{Quantitative results on generated samples from the MUG facial dataset. "DS-VAE(NA)" means that number of actions is not incorporated~\citep{li2018disentangled}. In "DS-VAE(NA)", samples are generated by fixing the encoded motions and randomly sampling content variable from the prior. Samples on DS-VAE(W), R-WAE(MMD) and R-WAE(GAN) are generated by incorporating the prior information(number of actions) into the model. \label{quant:mug}} 
 \end{center}
 \end{table*}
 \vspace{-0.4cm}
% \begin{table*}
% \begin{center}
% \resizebox{1\textwidth}{!}{
%   \begin{tabular}{l|ccccc} \hline
%  Metrics &{\small MocoGAN} & {\small DS-VAE(NA)}& {\small DS-VAE(W)} &{\small R-WAE(MMD)}&{\small R-WAE(GAN)}\\ \hline
%  {\small Accuracy ~$\uparrow$} & 75.50\%  & 66.73\% & 82.84 \% & 88.62\% & {\bf 90.15\%}   \\ %\hline
%  {\small Intra-entropy$\downarrow$} & 0.26 & 0.28& 0.23 & 0.17 & {\bf 0.15} \\ %\hline
% {\small Inter-entropy$\uparrow$} & 1.78 & 1.77 & 1.78 & {\bf 1.79} & {\bf 1.79}\\ %\hline
% {\small Inception Score~$\uparrow$} & 4.60 & 4.44 & 4.71 & 5.05 & {\bf 5.16}  \\ %\hline
% {\small ~~~ FID~$\downarrow$} & 16.95& 18.72&14.79& {\bf 12.21}& {\bf 10.86} \\
% \hline
%  \end{tabular}}
% \vspace{-0.3cm}
%  \caption{Quantitative results on generated samples from the MUG facial dataset. "DS-VAE(NA)" means that number of actions is not incorporated~\cite{li2018disentangled}. In "DS-VAE(NA)", samples are generated by fixing the encoded motions and randomly sampling content variable from the prior. Samples on DS-VAE(W), R-WAE(MMD) and R-WAE(GAN) are generated by incorporating the prior information(number of actions) into the model. \label{quant:mug}} 
%  \end{center}
%  \end{table*}
\vspace{-.1cm}
\paragraph{MUG Facial Dataset} We quantitatively evaluate the disentanglement and quality of generated samples. We train a 3D classifier on MUG facial dataset with accuracy 95.11\% and Inception Score 5.20 on test data~\citep{salimans2016improved}. We calculate Inception score, intra-entropy $H(y|\bd{v})$, where $y$ is the predicted label and $\bd{v}$ is the generated video, and inter-entropy $H(y)$~\citep{he2018probabilistic}. For a comprehensive quantitative evaluation, Frame-level FID score, introduced by~\citep{heusel2017gans}, is also provided. From Table 2, our R-WAE(MMD) and R-WAE(GAN) have higher accuracy, which means the categorical variable best captures the actions, which indicates our models achieve the best disentanglement. In table 2, the Inception score of R-WAE(GAN) is very close to Inception Score of the exact test data, which means our models have the best sample quality. Our proposed R-WAE(GAN) and R-WAE(MMD) have the best frame-level FID scores, compared with DS-VAE and MoCoGAN. The orignal DS-VAE (DS-VAE(NA))~\citep{li2018disentangled} without leveraging the number of actions performs worst, which shows that incorporating the number of actions as prior information does enhance the disentanglement of actions. % \textbf{Erran: the following sentence does not parse; needs to be rewritten} 

% \subsection{Results on MUG Facial Dataset}
%\paragraph{Results on MUG Facial Dataset} Like MoCoGAN~\cite{tulyakov2018mocogan}, we leverage the prior information (number of actions is 6) to reinforce disentanglement. We incorporate a latent action variable $\va$ (a one-hot vector with dimension 6) into our model to enhance disentanglement. The regularization for learning $\va$ is described in Section \ref{mainmethod}. The latent motion variable is augmented as $\vz^R_t=(\va, \vz^m_t).$
\vspace{-0.1cm}
\subsection{Unconditional Video Generation}
\vspace{-0.1cm}
%\textbf{erran: do we need the following text somewhere?
% seems ok to cut.
%}
%MoCoGAN~\cite{tulyakov2018mocogan} is not tailored for SM-MNIST dataset since their training requires finite number of actions in order to disentangle the action. The fact that MoCoGAN doesn't perform well on SM-MNIST can be easily seen from Fig.~\ref{fig:mnistsample}. We find that the performance of MoCoGAN~\cite{tulyakov2018mocogan} is much worse than DS-VAE~\cite{li2018disentangled} and our R-WAE(MMD) on SM-MNIST and Sprite. On MUG dataset, our model can swap the smile and fear action between two persons.   

%input{tex/mnist_video.tex}
\paragraph{SM-MNIST dataset} Fig.~\ref{fig:mnistsample} in Appendix E provides generated samples on the SM-MNIST dataset by randomly sampling content $\{\vz^c\}$ from the prior $p(\vz^c)$ and motions $\{\vz^m_{1:T}\}$ from the learned prior $p_\psi(\vz_t^m|\vz_{<t}^m).$ The length of our generated videos is $T=100$ and we only show randomly chosen videos of $T=20$ to save file size. Our R-WAE(MMD) achieves the most consistent and visually best sequence even when $T=100$. Samples from MoCoGAN~\citep{tulyakov2018mocogan} usually change digit identity along the sequence. The reason is that MoCoGAN~\citep{tulyakov2018mocogan} requires the number of actions be finite. Our generated Sprites videos also have the best results but are not provided due to  page limit.  
\vspace{-0.1cm}
 \begin{figure*}[ht]
\centering
\begin{subfigure}{0.33333\textwidth}
\centering
\begingroup
\begin{tabular}{cccc}
\includegraphics[height=.23\textwidth]{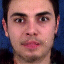} & \hspace{-0.5cm}
\includegraphics[height=.23\textwidth]{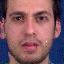} & \hspace{-0.5cm}
\includegraphics[height=.23\textwidth]{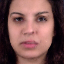} & \hspace{-0.5cm}
\includegraphics[height=.23\textwidth]{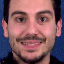} \\
\includegraphics[height=.23\textwidth]{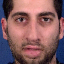} & \hspace{-0.5cm}
\includegraphics[height=.23\textwidth]{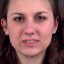} & \hspace{-0.5cm}
\includegraphics[height=.23\textwidth]{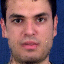} & \hspace{-0.5cm}
\includegraphics[height=.23\textwidth]{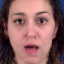}  \\
\includegraphics[height=.23\textwidth]{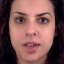} & \hspace{-0.5cm}
\includegraphics[height=.23\textwidth]{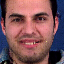} & \hspace{-0.5cm}
\includegraphics[height=.23\textwidth]{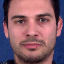} & \hspace{-0.5cm}
\includegraphics[height=.23\textwidth]{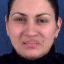}  \\
\includegraphics[height=.23\textwidth]{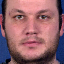} & \hspace{-0.5cm}
\includegraphics[height=.23\textwidth]{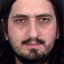} & \hspace{-0.5cm}
\includegraphics[height=.23\textwidth]{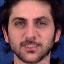} & \hspace{-0.5cm}
\includegraphics[height=.23\textwidth]{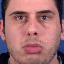}  \\
\includegraphics[height=.23\textwidth]{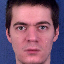} & \hspace{-0.5cm}
\includegraphics[height=.23\textwidth]{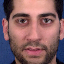} & \hspace{-0.5cm}
\includegraphics[height=.23\textwidth]{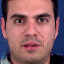} & \hspace{-0.5cm}
\includegraphics[height=.23\textwidth]{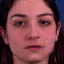}   \\
\end{tabular}
\endgroup
\vspace{-0.2cm}
\caption{R-WAE(GAN)}
\end{subfigure}%
\hspace{-.3cm}
\begin{subfigure}{0.33333\textwidth}
\centering
\begin{tabular}{cccc}
\includegraphics[height=.23\textwidth]{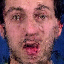} & \hspace{-0.5cm}
\includegraphics[height=.23\textwidth]{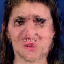} & \hspace{-0.5cm}
\includegraphics[height=.23\textwidth]{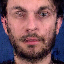} & \hspace{-0.5cm}
\includegraphics[height=.23\textwidth]{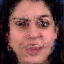} \\
\includegraphics[height=.23\textwidth]{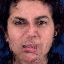} & \hspace{-0.5cm}
\includegraphics[height=.23\textwidth]{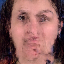} & \hspace{-0.5cm}
\includegraphics[height=.23\textwidth]{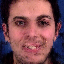} & \hspace{-0.5cm}
\includegraphics[height=.23\textwidth]{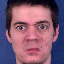} \\
\includegraphics[height=.23\textwidth]{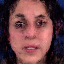} & \hspace{-0.5cm}
\includegraphics[height=.23\textwidth]{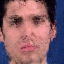} & \hspace{-0.5cm}
\includegraphics[height=.23\textwidth]{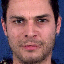} & \hspace{-0.5cm}
\includegraphics[height=.23\textwidth]{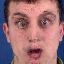} \\
\includegraphics[height=.23\textwidth]{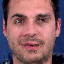} & \hspace{-0.5cm} 
\includegraphics[height=.23\textwidth]{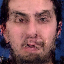} & \hspace{-0.5cm}
\includegraphics[height=.23\textwidth]{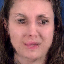} & \hspace{-0.5cm}
\includegraphics[height=.23\textwidth]{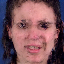} \\
\includegraphics[height=.23\textwidth]{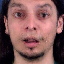} & \hspace{-0.5cm}
\includegraphics[height=.23\textwidth]{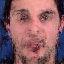} & \hspace{-0.5cm}
\includegraphics[height=.23\textwidth]{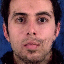} & \hspace{-0.5cm}
\includegraphics[height=.23\textwidth]{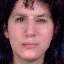} \\
\end{tabular}
\vspace{-0.2cm}
\caption{DS-VAE}
\end{subfigure}
\hspace{-.3cm}
\begin{subfigure}{0.33333\textwidth}
\centering
\label{fig:mug_mocogan}
\begingroup
\begin{tabular}{cccc}
\includegraphics[height=.23\textwidth]{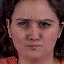} & \hspace{-0.5cm}
\includegraphics[height=.23\textwidth]{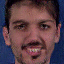} & \hspace{-0.5cm}
\includegraphics[height=.23\textwidth]{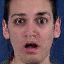} & \hspace{-0.5cm}
\includegraphics[height=.23\textwidth]{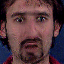} \\
\includegraphics[height=.23\textwidth]{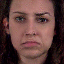} & \hspace{-0.5cm}
\includegraphics[height=.23\textwidth]{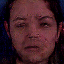} & \hspace{-0.5cm}
\includegraphics[height=.23\textwidth]{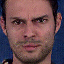} & \hspace{-0.5cm}
\includegraphics[height=.23\textwidth]{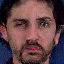} \\
\includegraphics[height=.23\textwidth]{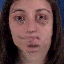} & \hspace{-0.5cm}
\includegraphics[height=.23\textwidth]{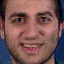} & \hspace{-0.5cm}
\includegraphics[height=.23\textwidth]{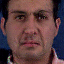} & \hspace{-0.5cm}
\includegraphics[height=.23\textwidth]{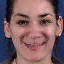} \\
\includegraphics[height=.23\textwidth]{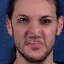} & \hspace{-0.5cm}
\includegraphics[height=.23\textwidth]{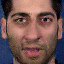} & \hspace{-0.5cm}
\includegraphics[height=.23\textwidth]{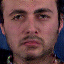} & \hspace{-0.5cm}
\includegraphics[height=.23\textwidth]{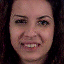} \\
\includegraphics[height=.23\textwidth]{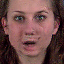} & \hspace{-0.5cm}
\includegraphics[height=.23\textwidth]{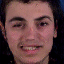} & \hspace{-0.5cm}
\includegraphics[height=.23\textwidth]{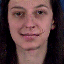} & \hspace{-0.5cm}
\includegraphics[height=.23\textwidth]{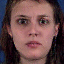} \\
\end{tabular}
\endgroup 
\vspace{-0.2cm}
\caption{MoCoGAN}
\end{subfigure}
\vspace{-0.2cm}
\caption{Unconditional video generation on MUG dataset, where the sample at time step $T=10$ is chosen for clear comparison. DS-VAE in (b) is improved by incorporating categorical latent variables. Samples of the video sequence  are given in Appendix E.\label{fig:mugframe}} 
\end{figure*}

\vspace{-0.4cm}
  
% \paragraph{Qualitative Results} we reconstruct videos by swapped encoded codes $\{\vz^c, \vz_t^R\}$ of video $A$ and $B$ on test data in Fig.~\ref{fig:dismug}. We randomly choose two sets of video sequences to illustrate the disentanglement. In Fig.~\ref{fig:dismug} (left), the identity changes from the fear expression to the happy expression or reverse. In Fig.~\ref{fig:dismug} (right), the identity changes from the happy expression to the anger expression or reverse. 

 \vspace{-0.1cm}
\paragraph{MUG Facial Dataset} Fig.~\ref{fig:mugframe} and Fig.~\ref{fig:mugsample} in Appendix E show generated samples on MUG dataset by randomly sampling content $\{\vz^c\}$ from the prior $p(\vz^c)$ and motions $\vz^R_t=(\va, \vz_t^m)$ from the categorical prior $p(\va)$ (latent action variable $\va$ is a one-hot vector with dimension 6) and the learned prior $p_{\psi}(\vz_t^m|\vz_{<t}^m).$ We show generated videos of length $T=10$. DS-VAE~\citep{li2018disentangled} used the same structure as ours. Fig.~\ref{fig:mugsample} shows that DS-VAE~\citep{li2018disentangled} and MoCoGAN~\citep{tulyakov2018mocogan} have blurry beginning frames $\{\vx_t\}$ and even more blurry frames as time $t$ evolves. While our R-WAE(GAN) has much better frame quality and more consistent video sequences. To have a clear comparison among all three methods, we show the samples at time step $T=10$ in Fig.~\ref{fig:mugframe}, and we can see that DS-VAE has very blurry samples with large time steps.  

% \begin{figure}[tbh]
% \centering
% \begingroup
% \setlength{\tabcolsep}{.3pt}
% \begin{tabular}{cc}
%  \multicolumn{2}{c}{{\small R-WAE~(MMD)}} \vspace{-.1cm}\\ \vspace{-.15cm} 
% \includegraphics[height=0.199\textwidth]{figures/latent_wae.pdf}& 
% \includegraphics[height=0.199\textwidth]{figures/latent_wae_long.pdf} \\
% {\small (a) 100 Epochs} & {\small (b) 500 Epochs} \\
% \end{tabular}
% \begin{tabular}{cc}
% \multicolumn{2}{c}{{\small Li\&Mandt, 2018}} \vspace{-.1cm} \\ \vspace{-.15cm} 
% \includegraphics[height=0.199\textwidth]{figures/latent_kl.pdf} &
% \includegraphics[height=0.199\textwidth]{figures/latent_kl_long.pdf}\\
% {\small (c) 100 Epochs} & {\small (d) 500 Epochs} \\
% \end{tabular}
% \endgroup
% \caption{\small Visualization of latent manifolds on SM-MNIST dataset. The static parts of both methods can learn the invariant digit information in the beginning training epochs. It is interesting to observe that Li\&Mandt tends to collapse after long training while ours are stable. \label{fig:latent}}
% \end{figure}

%erran: \section{Conclusion and Discussion}
\section{Conclusion}
%erran: below is too detailed for a conclusion. Typically there is no need to mention dsatasets in conclusion section.
%We propose a recurrent Wasserstein autoencoder framework to learn the disentangled representations of sequential data based on the optimal transport for distributions with sequential variables. Results on SM-MNIST, Sprites and TIMIT speech datasets demonstrate that our algorithm can effectively disentangle the static and dynamic latent factors of input sequences in a completely unsupervised manner. 
%On SM-MNIST with stochastic motion, our model can disentangle the digit attribute and the motion exactly and unconditionally generate long sequence. We simultaneously demonstrates that inductive bias (a two-flow model) is necessary to the success of disentangled representation representation. On Sprites where the attributes of characters are independent, our model can disentangle these factors within the static latent code.  By incorporating an additional categorical latent variable on MUG facial dataset, our model achieves the states-of-the-art results on the disentanglement of the invariant and variant latent information and unconditional video generation under the same setting as baselines. Future research directions include exploring our framework in conditional settings for text-to-video synthesis.
%
In this paper, we propose recurrent Wasserstein Autoencoder (R-WAE) %framework 
to learn disentangled representations of sequential data based on the optimal transport between distributions with sequential variables. Our theoretical analysis shows that R-WAE simultaneously maximizes the mutual information between input sequential data and different disentangled latent factors. Experiments on a variety of datasets demonstrate that our models achieve state-of-the-art results on the disentanglement of static and dynamic latent representations and unconditional video generation. Future research includes exploring our framework in self-supervised learning and conditional settings for text-to-video and video-to-video synthesis.
\paragraph{Acknowledgement} Jun Han thanks insightful discussions from Dr. Chen Fang at Tencent and invaluable support from Prof. Qiang Liu at UT Austin.
%\newpage
%\section*{Broader Impact}
%In order to provide a balanced perspective, authors are required to include a statement of the potential broader impact of their work, including its ethical aspects and future societal consequences. Authors should take care to discuss both positive and negative outcomes.
%Our work is the first to explore the Wasserstein metric for sequential representation learning and set the new state-of-the-art for unsupervised representation disentanglement of sequential data. We expect that this foundational framework will be further studied and widely applicable in many applications. It can be used for unconditional video generation in movie and gaming industry, anonymizing speaker identity in automatic speech recognition, and generating synthetic videos for education and training self-driving cars. However, it also has the risk of being abused. For example, it can be used to generate fake videos that look realistic. 
{\small
\bibliographystyle{iclr2020_conference}
\bibliography{rwvae}

\begin{thebibliography}{57}
\providecommand{\natexlab}[1]{#1}
\providecommand{\url}[1]{\texttt{#1}}
\expandafter\ifx\csname urlstyle\endcsname\relax
  \providecommand{\doi}[1]{doi: #1}\else
  \providecommand{\doi}{doi: \begingroup \urlstyle{rm}\Url}\fi

\bibitem[Aifanti et~al.(2010)Aifanti, Papachristou, and
  Delopoulos]{aifanti2010mug}
Niki Aifanti, Christos Papachristou, and Anastasios Delopoulos.
\newblock The mug facial expression database.
\newblock In \emph{11th International Workshop on Image Analysis for Multimedia
  Interactive Services WIAMIS 10}. IEEE, 2010.

\bibitem[Arbel et~al.(2018)Arbel, Sutherland, Bi{\'n}kowski, and
  Gretton]{arbel2018gradient}
Michael Arbel, Dougal Sutherland, Miko{\l}aj Bi{\'n}kowski, and Arthur Gretton.
\newblock On gradient regularizers for mmd gans.
\newblock In \emph{NIPS}, 2018.

\bibitem[Arjovsky et~al.(2018)Arjovsky, Chintala, and
  Bottou]{arjovsky2017wasserstein}
Martin Arjovsky, Soumith Chintala, and L{\'e}on Bottou.
\newblock Wasserstein gan.
\newblock \emph{ICML}, 2018.

\bibitem[Balaji et~al.(2018)Balaji, Min, Bai, Chellappa, and
  Graf]{balaji2018tfgan}
Yogesh Balaji, Martin~Renqiang Min, Bing Bai, Rama Chellappa, and Hans~Peter
  Graf.
\newblock Tfgan: Improving conditioning for text-to-video synthesis.
\newblock 2018.

\bibitem[Bellemare et~al.(2017)Bellemare, Danihelka, Dabney, Mohamed,
  Lakshminarayanan, Hoyer, and Munos]{bellemare2017cramer}
Marc~G Bellemare, Ivo Danihelka, Will Dabney, Shakir Mohamed, Balaji
  Lakshminarayanan, Stephan Hoyer, and R{\'e}mi Munos.
\newblock The cramer distance as a solution to biased wasserstein gradients.
\newblock \emph{arXiv preprint arXiv:1705.10743}, 2017.

\bibitem[Bengio et~al.(2013)Bengio, Courville, and
  Vincent]{bengio2013representation}
Yoshua Bengio, Aaron Courville, and Pascal Vincent.
\newblock Representation learning: A review and new perspectives.
\newblock \emph{IEEE transactions on pattern analysis and machine
  intelligence}, 2013.

\bibitem[Bi{\'n}kowski et~al.(2018)Bi{\'n}kowski, Sutherland, Arbel, and
  Gretton]{binkowski2018demystifying}
Miko{\l}aj Bi{\'n}kowski, Dougal~J Sutherland, Michael Arbel, and Arthur
  Gretton.
\newblock Demystifying mmd gans.
\newblock \emph{arXiv preprint arXiv:1801.01401}, 2018.

\bibitem[Bousquet et~al.(2017)Bousquet, Gelly, Tolstikhin, Simon-Gabriel, and
  Schoelkopf]{bousquet2017optimal}
Olivier Bousquet, Sylvain Gelly, Ilya Tolstikhin, Carl-Johann Simon-Gabriel,
  and Bernhard Schoelkopf.
\newblock From optimal transport to generative modeling: the vegan cookbook.
\newblock \emph{arXiv preprint arXiv:1705.07642}, 2017.

\bibitem[Brock et~al.(2019)Brock, Donahue, and Simonyan]{brock2018large}
Andrew Brock, Jeff Donahue, and Karen Simonyan.
\newblock Large scale gan training for high fidelity natural image synthesis.
\newblock \emph{ICLR}, 2019.

\bibitem[Chen et~al.(2018)Chen, Li, Grosse, and Duvenaud]{chen2018isolating}
Tian~Qi Chen, Xuechen Li, Roger~B Grosse, and David~K Duvenaud.
\newblock Isolating sources of disentanglement in variational autoencoders.
\newblock In \emph{NeurIPS}, 2018.

\bibitem[Chen et~al.(2016)Chen, Duan, Houthooft, Schulman, Sutskever, and
  Abbeel]{chen2016infogan}
Xi~Chen, Yan Duan, Rein Houthooft, John Schulman, Ilya Sutskever, and Pieter
  Abbeel.
\newblock Infogan: Interpretable representation learning by information
  maximizing generative adversarial nets.
\newblock In \emph{NIPS}, 2016.

\bibitem[Chung et~al.(2015)Chung, Kastner, Dinh, Goel, Courville, and
  Bengio]{vrnn_nips2015}
Junyoung Chung, Kyle Kastner, Laurent Dinh, Kratarth Goel, Aaron~C Courville,
  and Yoshua Bengio.
\newblock A recurrent latent variable model for sequential data.
\newblock In C.~Cortes, N.~D. Lawrence, D.~D. Lee, M.~Sugiyama, and R.~Garnett
  (eds.), \emph{Advances in Neural Information Processing Systems 28}, pp.\
  2980--2988. Curran Associates, Inc., 2015.

\bibitem[Dehak et~al.(2010)Dehak, Kenny, Dehak, Dumouchel, and
  Ouellet]{dehak2010front}
Najim Dehak, Patrick~J Kenny, R{\'e}da Dehak, Pierre Dumouchel, and Pierre
  Ouellet.
\newblock Front-end factor analysis for speaker verification.
\newblock \emph{IEEE Transactions on Audio, Speech, and Language Processing},
  2010.

\bibitem[Denton et~al.(2017)]{denton2017unsupervised}
Emily~L Denton et~al.
\newblock Unsupervised learning of disentangled representations from video.
\newblock In \emph{NIPS}, 2017.

\bibitem[Dupont(2018)]{dupont2018learning}
Emilien Dupont.
\newblock Learning disentangled joint continuous and discrete representations.
\newblock In \emph{NIPS}, 2018.

\bibitem[Esfahani \& Kuhn(2018)Esfahani and Kuhn]{esfahani2018data}
Peyman~Mohajerin Esfahani and Daniel Kuhn.
\newblock Data-driven distributionally robust optimization using the
  wasserstein metric: Performance guarantees and tractable reformulations.
\newblock \emph{Mathematical Programming}, 2018.

\bibitem[Garofolo(1993)]{garofolo1993timit}
John~S Garofolo.
\newblock Timit acoustic phonetic continuous speech corpus.
\newblock \emph{Linguistic Data Consortium, 1993}, 1993.

\bibitem[Goodfellow et~al.(2014)Goodfellow, Pouget-Abadie, Mirza, Xu,
  Warde-Farley, Ozair, Courville, and Bengio]{goodfellow2014generative}
Ian Goodfellow, Jean Pouget-Abadie, Mehdi Mirza, Bing Xu, David Warde-Farley,
  Sherjil Ozair, Aaron Courville, and Yoshua Bengio.
\newblock Generative adversarial nets.
\newblock In \emph{NIPS}, 2014.

\bibitem[Gretton et~al.(2007)Gretton, Borgwardt, Rasch, Sch{\"o}lkopf, and
  Smola]{gretton2007kernel}
Arthur Gretton, Karsten Borgwardt, Malte Rasch, Bernhard Sch{\"o}lkopf, and
  Alex~J Smola.
\newblock A kernel method for the two-sample-problem.
\newblock In \emph{NIPS}, 2007.

\bibitem[Gulrajani et~al.(2017)Gulrajani, Ahmed, Arjovsky, Dumoulin, and
  Courville]{gulrajani2017improved}
Ishaan Gulrajani, Faruk Ahmed, Martin Arjovsky, Vincent Dumoulin, and Aaron~C
  Courville.
\newblock Improved training of wasserstein gans.
\newblock In \emph{NIPS}, 2017.

\bibitem[Han et~al.(2018)Han, Lombardo, Schroers, and Mandt]{han2018deep}
Jun Han, Salvator Lombardo, Christopher Schroers, and Stephan Mandt.
\newblock Deep probabilistic video compression.
\newblock \emph{arXiv preprint arXiv:1810.02845}, 2018.

\bibitem[He et~al.(2018)He, Lehrmann, Marino, Mori, and
  Sigal]{he2018probabilistic}
Jiawei He, Andreas Lehrmann, Joseph Marino, Greg Mori, and Leonid Sigal.
\newblock Probabilistic video generation using holistic attribute control.
\newblock In \emph{Proceedings of the European Conference on Computer Vision
  (ECCV)}, pp.\  452--467, 2018.

\bibitem[He et~al.(2019)He, Spokoyny, Neubig, and
  Berg-Kirkpatrick]{he2019lagging}
Junxian He, Daniel Spokoyny, Graham Neubig, and Taylor Berg-Kirkpatrick.
\newblock Lagging inference networks and posterior collapse in variational
  autoencoders.
\newblock \emph{ICLR}, 2019.

\bibitem[He et~al.(2016)He, Zhang, Ren, and Sun]{he2016deep}
Kaiming He, Xiangyu Zhang, Shaoqing Ren, and Jian Sun.
\newblock Deep residual learning for image recognition.
\newblock In \emph{CVPR}, 2016.

\bibitem[Heusel et~al.(2017)Heusel, Ramsauer, Unterthiner, Nessler, and
  Hochreiter]{heusel2017gans}
Martin Heusel, Hubert Ramsauer, Thomas Unterthiner, Bernhard Nessler, and Sepp
  Hochreiter.
\newblock Gans trained by a two time-scale update rule converge to a local nash
  equilibrium.
\newblock In \emph{NeurIPS}, 2017.

\bibitem[Higgins et~al.(2017)Higgins, Matthey, Pal, Burgess, Glorot, Botvinick,
  Mohamed, and Lerchner]{higgins2017beta}
Irina Higgins, Loic Matthey, Arka Pal, Christopher Burgess, Xavier Glorot,
  Matthew Botvinick, Shakir Mohamed, and Alexander Lerchner.
\newblock beta-vae: Learning basic visual concepts with a constrained
  variational framework.
\newblock In \emph{ICLR}, 2017.

\bibitem[Hjelm et~al.(2018)Hjelm, Fedorov, Lavoie-Marchildon, Grewal,
  Trischler, and Bengio]{hjelm2018learning}
R~Devon Hjelm, Alex Fedorov, Samuel Lavoie-Marchildon, Karan Grewal, Adam
  Trischler, and Yoshua Bengio.
\newblock Learning deep representations by mutual information estimation and
  maximization.
\newblock \emph{arXiv preprint arXiv:1808.06670}, 2018.

\bibitem[Hsieh et~al.(2018)Hsieh, Liu, Huang, Fei-Fei, and
  Niebles]{hsieh2018learning}
Jun-Ting Hsieh, Bingbin Liu, De-An Huang, Li~F Fei-Fei, and Juan~Carlos
  Niebles.
\newblock Learning to decompose and disentangle representations for video
  prediction.
\newblock In \emph{NeurIPS}, 2018.

\bibitem[Hsu et~al.(2017)Hsu, Zhang, and Glass]{hsu2017unsupervised}
Wei-Ning Hsu, Yu~Zhang, and James Glass.
\newblock Unsupervised learning of disentangled and interpretable
  representations from sequential data.
\newblock In \emph{NIPS}, 2017.

\bibitem[Jang et~al.(2016)Jang, Gu, and Poole]{jang2016categorical}
Eric Jang, Shixiang Gu, and Ben Poole.
\newblock Categorical reparameterization with gumbel-softmax.
\newblock 2016.

\bibitem[Kim \& Mnih(2018)Kim and Mnih]{kim2018disentangling}
Hyunjik Kim and Andriy Mnih.
\newblock Disentangling by factorising.
\newblock In \emph{ICML}, 2018.

\bibitem[Kingma \& Ba(2015)Kingma and Ba]{kingma2014adam}
Diederik~P Kingma and Jimmy Ba.
\newblock Adam: A method for stochastic optimization.
\newblock \emph{ICLR}, 2015.

\bibitem[Kingma \& Welling(2014)Kingma and Welling]{kingma2013auto}
Diederik~P Kingma and Max Welling.
\newblock Auto-encoding variational bayes.
\newblock \emph{NIPS}, 2014.

\bibitem[Li et~al.(2017)Li, Chang, Cheng, Yang, and P{\'o}czos]{li2017mmd}
Chun-Liang Li, Wei-Cheng Chang, Yu~Cheng, Yiming Yang, and Barnab{\'a}s
  P{\'o}czos.
\newblock Mmd gan: Towards deeper understanding of moment matching network.
\newblock In \emph{NIPS}, 2017.

\bibitem[Locatello et~al.(2019)Locatello, Bauer, Lucic, Raetsch, Gelly,
  Sch{\"o}lkopf, and Bachem]{locatello2019challenging}
Francesco Locatello, Stefan Bauer, Mario Lucic, Gunnar Raetsch, Sylvain Gelly,
  Bernhard Sch{\"o}lkopf, and Olivier Bachem.
\newblock Challenging common assumptions in the unsupervised learning of
  disentangled representations.
\newblock In \emph{International Conference on Machine Learning}, 2019.

\bibitem[Maaten \& Hinton(2008)Maaten and Hinton]{maaten2008visualizing}
Laurens van~der Maaten and Geoffrey Hinton.
\newblock Visualizing data using t-sne.
\newblock \emph{JMLR}, 2008.

\bibitem[Maddison et~al.(2016)Maddison, Mnih, and Teh]{maddison2016concrete}
Chris~J Maddison, Andriy Mnih, and Yee~Whye Teh.
\newblock The concrete distribution: A continuous relaxation of discrete random
  variables.
\newblock \emph{arXiv preprint arXiv:1611.00712}, 2016.

\bibitem[Makhzani et~al.(2015)Makhzani, Shlens, Jaitly, Goodfellow, and
  Frey]{makhzani2015adversarial}
Alireza Makhzani, Jonathon Shlens, Navdeep Jaitly, Ian Goodfellow, and Brendan
  Frey.
\newblock Adversarial autoencoders.
\newblock \emph{arXiv preprint arXiv:1511.05644}, 2015.

\bibitem[Mescheder et~al.(2018)Mescheder, Geiger, and
  Nowozin]{mescheder2018training}
Lars Mescheder, Andreas Geiger, and Sebastian Nowozin.
\newblock Which training methods for gans do actually converge?
\newblock \emph{ICML}, 2018.

\bibitem[Miyato et~al.(2018)Miyato, Kataoka, Koyama, and
  Yoshida]{miyato2018spectral}
Takeru Miyato, Toshiki Kataoka, Masanori Koyama, and Yuichi Yoshida.
\newblock Spectral normalization for generative adversarial networks.
\newblock \emph{ICLR}, 2018.

\bibitem[Nowozin et~al.(2016)Nowozin, Cseke, and Tomioka]{nowozin2016f}
Sebastian Nowozin, Botond Cseke, and Ryota Tomioka.
\newblock f-gan: Training generative neural samplers using variational
  divergence minimization.
\newblock In \emph{NIPS}, pp.\  271--279, 2016.

\bibitem[Ozair et~al.(2019)Ozair, Lynch, Bengio, Van~den Oord, Levine, and
  Sermanet]{ozair2019wasserstein}
Sherjil Ozair, Corey Lynch, Yoshua Bengio, Aaron Van~den Oord, Sergey Levine,
  and Pierre Sermanet.
\newblock Wasserstein dependency measure for representation learning.
\newblock In \emph{Advances in Neural Information Processing Systems}, 2019.

\bibitem[Petzka et~al.(2017)Petzka, Fischer, and
  Lukovnicov]{petzka2017regularization}
Henning Petzka, Asja Fischer, and Denis Lukovnicov.
\newblock On the regularization of wasserstein gans.
\newblock \emph{NIPS}, 2017.

\bibitem[Poole et~al.(2019)Poole, Ozair, Van Den~Oord, Alemi, and
  Tucker]{poole2019variational}
Ben Poole, Sherjil Ozair, Aaron Van Den~Oord, Alex Alemi, and George Tucker.
\newblock On variational bounds of mutual information.
\newblock In \emph{ICML}, 2019.

\bibitem[Qi(2017)]{qi2017loss}
Guo-Jun Qi.
\newblock Loss-sensitive generative adversarial networks on lipschitz
  densities.
\newblock \emph{arXiv preprint arXiv:1701.06264}, 2017.

\bibitem[Roth et~al.(2017)Roth, Lucchi, Nowozin, and
  Hofmann]{roth2017stabilizing}
Kevin Roth, Aurelien Lucchi, Sebastian Nowozin, and Thomas Hofmann.
\newblock Stabilizing training of generative adversarial networks through
  regularization.
\newblock In \emph{NIPS}, 2017.

\bibitem[Rubenstein et~al.(2018{\natexlab{a}})Rubenstein, Schoelkopf, and
  Tolstikhin]{rubenstein2018latent}
Paul~K Rubenstein, Bernhard Schoelkopf, and Ilya Tolstikhin.
\newblock On the latent space of wasserstein auto-encoders.
\newblock \emph{arXiv preprint arXiv:1802.03761}, 2018{\natexlab{a}}.

\bibitem[Rubenstein et~al.(2018{\natexlab{b}})Rubenstein, Schoelkopf, and
  Tolstikhin]{rubenstein2018learning}
Paul~K Rubenstein, Bernhard Schoelkopf, and Ilya Tolstikhin.
\newblock Learning disentangled representations with wasserstein auto-encoders.
\newblock 2018{\natexlab{b}}.

\bibitem[Salimans et~al.(2016)Salimans, Goodfellow, Zaremba, Cheung, Radford,
  and Chen]{salimans2016improved}
Tim Salimans, Ian Goodfellow, Wojciech Zaremba, Vicki Cheung, Alec Radford, and
  Xi~Chen.
\newblock Improved techniques for training gans.
\newblock In \emph{NIPS}, pp.\  2234--2242, 2016.

\bibitem[S{\o}nderby et~al.(2016)S{\o}nderby, Raiko, Maal{\o}e, S{\o}nderby,
  and Winther]{sonderby2016ladder}
Casper~Kaae S{\o}nderby, Tapani Raiko, Lars Maal{\o}e, S{\o}ren~Kaae
  S{\o}nderby, and Ole Winther.
\newblock Ladder variational autoencoders.
\newblock In \emph{NIPS}, 2016.

\bibitem[Sun et~al.(2018)Sun, Xu, and Saenko]{sun2018two}
Ximeng Sun, Huijuan Xu, and Kate Saenko.
\newblock A two-stream variational adversarial network for video generation.
\newblock \emph{arXiv preprint arXiv:1812.01037}, 2018.

\bibitem[Thanh-Tung et~al.(2019)Thanh-Tung, Tran, and
  Venkatesh]{thanh2019improving}
Hoang Thanh-Tung, Truyen Tran, and Svetha Venkatesh.
\newblock Improving generalization and stability of generative adversarial
  networks.
\newblock \emph{ICLR}, 2019.

\bibitem[Tolstikhin et~al.(2018)Tolstikhin, Bousquet, Gelly, and
  Schoelkopf]{tolstikhin2017wasserstein}
Ilya Tolstikhin, Olivier Bousquet, Sylvain Gelly, and Bernhard Schoelkopf.
\newblock Wasserstein auto-encoders.
\newblock \emph{ICLR}, 2018.

\bibitem[Tschannen et~al.(2020)Tschannen, Djolonga, Rubenstein, Gelly, and
  Lucic]{tschannen2019mutual}
Michael Tschannen, Josip Djolonga, Paul~K Rubenstein, Sylvain Gelly, and Mario
  Lucic.
\newblock On mutual information maximization for representation learning.
\newblock \emph{ICLR}, 2020.

\bibitem[Tulyakov et~al.(2018)Tulyakov, Liu, Yang, and
  Kautz]{tulyakov2018mocogan}
Sergey Tulyakov, Ming-Yu Liu, Xiaodong Yang, and Jan Kautz.
\newblock Mocogan: Decomposing motion and content for video generation.
\newblock In \emph{CVPR}, 2018.

\bibitem[Yingzhen \& Mandt(2018)Yingzhen and Mandt]{li2018disentangled}
Li~Yingzhen and Stephan Mandt.
\newblock Disentangled sequential autoencoder.
\newblock In \emph{ICML}, 2018.

\bibitem[Zhao et~al.(2017)Zhao, Song, and Ermon]{zhao2017learning}
Shengjia Zhao, Jiaming Song, and Stefano Ermon.
\newblock Learning hierarchical features from deep generative models.
\newblock In \emph{ICML}, 2017.

\end{thebibliography}
}

\clearpage
\onecolumn
\section*{Appendix for Recurrent Wasserstein AutoEncoder}
%erran: theorem number does not match

%\textcolor{red}{ACs complain that the Proofs are   incorrectly labeled and we are careless about our paper}

% TensorFlow 1.12
\subsection*{Appendix A: Proof of Theorem 1}
In the following, we provide the proof of Theorem 1.
% \begin{thm*}
%\label{thm:t2}

{\bf Theorem 1}  For $P_G$ defined with deterministic $P_G(X|Z)$ and any function $Y=G(Z),$ 
\begin{equation}
W(P_\CD, P_G) =\inf_{Q: Q_{Z^c}= P_{Z^c}, Q_{Z_{1:T}^m }=P_{Z_{1:T}^m }}\sum_{t=1}^T \E_{P_\CD}\E_{Q(Z_t|X_t)}[c(X_t, G(Z_t))],     
\end{equation}
where $Q_{Z_{1:T}}$ is the marginal distribution of $Z_{1:T}$ when $X_{1:T}\sim P_\CD$ and $Z_{1:T}\sim Q(Z_{1:T}|X_{1:T})$ and $P_{Z_{1:T}}$ is the prior.
Based on the assumptions, the constraint set is equivalent to 
\begin{equation}
\label{app:ineq}
W(P_\CD, P_G) \le \inf_{Q\in \mathcal{S}} \sum_t \E_{P_\CD}\E_{Q(Z_t|X_t)}[c(X_t, G(Z_t))],    
\end{equation}
where the set $\mathcal{S}=\{Q: Q_{Z^c}= P_{Z^c}, Q_{Z_{1}^m}=P_{Z_{1}^m}, Q_{Z_{t}^m|Z_{<t}^m} = P_{Z_{t}^m|Z_{<t}^m}\}.$
% \end{thm*}

{\bf Proof:} Consider the sequential random variables $\CD=X_{1:T}$ and $Y_{1:T},$  the optimal transport between the distribution for  $\CD=X_{1:T}$ and the distribution for $Y_{1:T}$ induces a rich class of divergence,
\begin{equation}
W(P_\CD, P_G) := \inf_{\Gamma \sim \cp(X_{1:T}\sim P_\CD, Y_{1:T}\sim P_G)}\E_{(X_{1:T}, Y_{1:T})\sim \Gamma}[c(X_{1:T}, Y_{1:T})]   
\end{equation}
where $\cp(X_{1:T}\sim P_\CD, Y_{1:T}\sim P_G)$ is a set of all joint distributions of $(X_{1:T}, Y_{1:T})$ with marginals $P_\CD$ and $P_G$, respectively.

When we choose $c(\vx,\vy)=\|\vx-\vy\|^2,$ $c(X_{1:T}, Y_{1:T})=\sum_t \|X_t-Y_t\|^2$ by linearity. It is easy to derive the optimal transport for distributions with sequential random variables,
\begin{equation}
W(P_\CD, P_G) =\inf_{Q: Q_{Z_{1:T}}= P_{Z_{1:T}}}  \sum_t \E_{P_\CD}\E_{Q(Z_t|X_t)}[c(X_t, G(Z_t))].          
\end{equation}
Based on our assumption, the marginal distribution of the inference model satisfies
\begin{equation}
Q(Z_1, \cdots, Z_T) = Q(Z^c)Q(Z_1^m, \cdots,  Z_T^m)=Q(Z^c)\prod_t Q(Z^m_t|Z^m_{<t}).    
\end{equation}
The prior distribution $P(Z_1,\cdots, Z_T)$ satisfies
\begin{equation}
P(Z_1, \cdots, Z_T) = P(Z^c)P(Z_1^m, \cdots,  Z_T^m)=P(Z^c)\prod_t P(Z^m_t|Z^m_{<t}).    
\end{equation}
Since the set $\mathcal{S}$ is a subset of $\{Q: Q_{Z_{1:T}}= P_{Z_{1:T}}\}$, we derive the inequality \eqref{app:ineq}.

\subsection*{Appendix B: Proof of Theorem 2}
%erran: should be theorem 3
%In the following, we provide the proof of Theorem 2. 
In the following, we provide the proof of Theorem 2.
To make the notations easy to read, we use the density functions of corresponding distributions.

The joint generative distribution is
\begin{equation*}
p(\vx_{1:T}, \vz_{1:T})=p_{\psi}(\vz_{1:T})p_{\vthe}(\vx_{1:T}|\vz_{1:T}),    
\end{equation*}
where $p_{\psi}(\vz_{1:T})$ is the prior distribution and $p_{\vthe}(\vx_{1:T}|\vz_{1:T})$ is the decoder model.
And the corresponding joint inference distribution is 
$q_\ff(\vx_{1:T}, \vz_{1:T})=p_{\CD}(\vx_{1:T})q_\ff(\vz_{1:T}\mid\vx_{1:T}).$

If the MI between $\vz_{1:T}$ and $\vx_{1:T}$ is defined in terms of the inference  model $q$, we have the following lower bound with step-by-step derivations: 
\begin{align}
\label{appdef:mi}
 I(\vz_{1:T}; \vx_{1:T})\!&\! = \E_{q(\vx_{1:T},\vz_{1:T})}[\log \frac{ q_\ff(\vz_{1:T}|\vx_{1:T})}{ q_\ff (\vz_{1:T})}]  \\ \notag
 &\!\! = \E_{q(\vx_{1:T}, \vz_{1:T})}[\BD_{\KL}(q_\ff(\vz_{1:T}|\vx_{1:T}), p(\vz_{1:T}|\vx_{1:T}))\!+\!\! \log p(\vz_{1:T}|\vx_{1:T})\!-\!\log q_\ff (\vz_{1:T})] \\ \notag
 &\!\! \ge\E_{p_\CD}[\E_{q(\vz_{1:T}|\vx_{1:T})}[\log p(\vx_{1:T}|\vz_{1:T})+\log p(\vz_{1:T})- \log q_\ff (\vz_{1:T})\! - \!\log p(\CD)]]\\ \notag
 & \!\!\ge \sum_{t=1}^T   \E_{p(\mathcal{D})}\big[\E_{q_\ff(\vz_t|\vx_t)}[\log p_\vthe(\vx_t|\vz_t)]\big]\!\!-\!\E_{p(\mathcal{D})}[\E_{q_\ff(\vz_t|\vx_t)}[ \log q_{\ff}(\vz^c) \!\!-\!\! \log p(\vz^c)]\big] \\ \notag
 &-\!\!\sum_{t=1}^T \E_{p(\mathcal{D})}\big[\E_{q_\ff (\vz_t^m\!\mid\vx_t)}[ \log q_{\ff}(\vz_t^m|\vz_{<t}^m)- \log p(\vz_t^m|\vz_{<t}^m)+\log p(\CD)\big],  \notag 
\end{align}

where we use Bayesian's rule $p(\vz_{1:T}|\vx_{1:T})=\frac{p_\vthe(\vx_{1:T}|\vz_{1:T})p(\vz_{1:T})}{p(\CD)}.$
Maximizing the MI between $\vz_{1:T}$ and $\vx_{1:T}$ achieves state-of-the-art results in disentangled latent representation by using different regularizers for the static and dynamical latent variable with different priors~\citep{hjelm2018learning}. In practice, incorporating the mutual information $I(\vz^m_t, \vx_t)$ between element $\vx_t$ and motion $\vz^m_t$ might facilitate the disentanglement of the dynamical latent variable $\vz^m_t$.

{\bf Theorem 3} When its distribution divergence is chosen as $\KL$ divergence, the regularization terms in Eq.~\eqref{obj:loss} jointly minimize the $\KL$ divergence between the inference model $Q(Z_{1:T}|X_{1:T})$ and the prior model $P(Z_{1:T})$ and maximize the mutual information between $X_{1:T}$ and $Z_{1:T},$
\begin{align}
\label{eq:reg1}
\KL(Q(Z^c)||P(Z^c)) & = \E_{p_{\CD}}[\KL(Q(Z^c|X_{1:T})||P(Z^c))] - I(X_{1:T}; Z^c).  \nonumber   \notag \\
\KL(Q(Z_{t}^m|Z_{<t}^m)||P(Z_{t}^m|Z_{<t}^m)) & = 
\E_{p_{\CD}}[\KL(Q(Z_{t}^m|Z_{<t}^m, X_{1:T})||P(Z_{t}^m|Z_{<t}^m)] - I(X_{1:T}; Z_{t}^m|Z_{<t}^m). \notag  
\end{align}
{\bf Proof}: Denote $X_{\CD} = X_{1:T}$. As in the proof of Theorem 2, the mutual information between $Z_{1:T}$ and $X_{1:T}$ is defined in terms of the inference model $Q$, and we use the density functions of corresponding distributions to make the notations easy to read. Thus, 
\begin{equation*}
Q(Z_{1:T}) = E_{p_{\CD}} q(\vz_{1:T} | \vx_{1:T}).
\end{equation*}
According to the definition of mutual information, we have
\begin{align*}
I(X_{1:T}; Z^c) & = E_{p_{\CD}} \sum_{\vz^c}^{} p_{\CD}(\vx_{1:T}) q(\vz^c | \vx_{1:T}) \log \frac{p_{\CD}(\vx_{1:T})q(\vz^c | \vx_{1:T})}{p_{\CD}(\vx_{1:T}) q(\vz^c)} \\
                        & = E_{p_{\CD}} \sum_{\vz^c}^{} q(\vz^c | \vx_{1:T})  \log \frac{q(\vz^c | \vx_{1:T})}{q(\vz^c)} \\
                        & = E_{p_{\CD}} \sum_{\vz^c}^{} q(\vz^c | \vx_{1:T})  \log \frac{q(\vz^c | \vx_{1:T})}{p(\vz^c)} - E_{p_{\CD}} \sum_{\vz^c}^{} q(\vz^c | \vx_{1:T})  \log \frac{q(\vz^c)}{p(\vz^c)} \\
                        & = E_{p_{\CD}} \sum_{\vz^c}^{} q(\vz^c | \vx_{1:T})  \log \frac{q(\vz^c | \vx_{1:T})}{p(\vz^c)} - \sum_{\vz^c}^{} q(\vz^c)  \log \frac{q(\vz^c)}{p(\vz^c)} \\
                        & =  \E_{p_{\CD}}[\KL(Q(Z^c|X_{1:T})||P(Z^c))] - \KL(Q(Z^c)||P(Z^c)) \\
\end{align*}
Therefore,
\begin{equation*}
\KL(Q(Z^c)||P(Z^c)) = \E_{p_{\CD}}[\KL(Q(Z^c|X_{1:T})||P(Z^c))] - I(X_{1:T}; Z^c).
\end{equation*}
%\KL(Q(Z^c)||P(Z^c)) & = \E_{p_{\CD}}\E_{q(Z^C|\CD)}[\log \frac{q(\CD, Z^C)}{p(\CD, Z^C)}] - \E_{q(Z^C)}[\KL(q(\CD|Z^C)||P(\CD))] \notag \\ & = \E_{p_{\CD}}\E_{q(Z^C|\CD)}[\log \frac{q(Z^C|\CD)p(\CD)}{p(Z^C)p(\CD|Z^C)}] - I(Z^C;\CD) \notag \\
%& = \E_{p_{\CD}}[\KL(Q(Z^c|X_{1:T})||P(Z^c_{1:T}))] - I(Z^C;\CD)  -\E_{p_{\CD}}\E_{q(Z^C|\CD)}[\log p(\CD|Z^C)]. 

Similarly, we can prove the second equality in the theorem.
\subsection*{Appendix C: Datasets \label{app:data}}
{\bf Stochastic Moving MNIST(SM-MNIST) Dataset} Stochastic moving MNIST (SM-MNIST) consists of sequences of frames of size $64\times 64 \times 1$, containing one MNIST digit moving and bouncing
off edges of the frame (walls). We use one digit instead of two digits because two moving digits may collide, which changes the content of the dynamics and is inconsistent with our assumption. The digits in SM-MNIST move with a constant velocity along a trajectory until they hit at wall at which point they bounce off with a random speed and direction.

{\bf Sprites Dataset} We follow the same steps as in \cite{li2018disentangled} to process Sprites dataset, which consists of animated cartoon characters whose clothing, hairstyle, skin color and action can be fully controlled. We use 6 variants in each of 4 attribute categories (skin colors, tops, pants and hair style) and there are $6^4=1296$ unique characters in total, where 1000 of them are used for training and the rest of them are used for testing. We use 9 action categories (walking, casting spells and slashing, each with three different viewing angles.) The resulting dataset consists of video sequences with $T=8$ frames of size $64\times 64\times 3$.

{\bf MUG Facial Dataset} We use the MUG Facial Expression Database \citep{aifanti2010mug} for this experiment. The dataset consists
of 86 subjects. Each video consists of 50 to 160 frames.
To use the same network architecture for the whole video datasets in this paper,  we cropped the face regions and scaled to the same size $64\times 64 \times 3$. We use six facial expressions (anger, fear, disgust, happiness, sadness,  
and surprise). To ensure there is sufficient change in the facial expression along a video sequence, we choose every other frame in the original video sequences to form training and test video sequences of length $T=10.$ $80\%$ of the videos are used for training and $20\%$ of the videos are used for testing.

{\bf TIMIT Speech Dataset} The TIMIT dataset~\citep{garofolo1993timit} contains broadband 16k Hz of phonetically-balanced read speech. A total of 6300 utterances (5.4 hours) are presented with 10 sentences from each of 630 speakers. The data is preprocessed in the same way as in \citep{li2018disentangled} and \citep{hsu2017unsupervised}.  The raw speech waveforms are first split into sub-sequences of 200ms, and then preprocessed with sparse fast Fourier transform to obtain a 200 dimensional log-magnitude spectrum, computed every 10ms, i.e., we use $T=20$ for sequence $\vx_{1:T}.$ The dimension of $\vx_t$ is 200.

Now we explain the detail of the evaluation metric, equal error rate (EER), used on TIMIT dataset. Let $\vw^{\mathrm{test}}$ be the feature of test utterance $\vx_{1:T}^{\mathrm{test}}$ and $\vw^{\mathrm{target}}$ be the feature of test utterance $\vx_{1:T}^{\mathrm{target}}.$ The predicted identity is confirmed if the cosine similarity between $\vw^{\mathrm{test}}$ and $\vw^{\mathrm{target}}$, $\mathrm{cos}(\vw^{\mathrm{test}},\vw^{\mathrm{target}})$ is greater than some threshold $\epsilon$ used in \cite{dehak2010front}. The equal error rate (EER) means the false rejection
rate equals the false acceptance rate \citep{dehak2010front}. In the following, we will discuss the two choices of feature $\vw^{\mathrm{test}}$ for evaluations of all methods, 
$$\bd{\mu}^c=\frac{1}{N}\sum_{i=1}^N\E_{q(\vz^c|\vx^i_{1:T})}[\vz^c],$$
which is used to evaluate the disentanglement of $\vz^c;$
$$\bd{\mu}^m=\frac{1}{NT}\sum_{i=1}^N\sum_{j=1}^T\E_{q(\vz^m_t|\vx^i_{1:T})}[\vz^m_t],$$
which is used to evaluate the disentanglement of $\vz^m.$ For more details, please refer to \citep{dehak2010front,li2018disentangled, hsu2017unsupervised}. We use the same network architecture as in \cite{li2018disentangled} for a fair comparison on speech dataset. As the input dimension of speech is low, the encoder/decoder network is a 2-hidden-layer MLP with the hidden dimension $256.$ 

\subsection*{Appendix D: Choices of Regularizers}
In the following, we will discuss the choice of regularizers in R-WAE. To make notations easy to read, we use density functions for corresponding distributions. In both R-WAE(GAN) and R-WAE(MMD), we use the same regularizer for $\bbd(q(\vz_{t}^m|\vz_{<t}^m), p(\vz_{t}^m|\vz_{<t}^m)).$ In experiments, we assume inference model $q(\vz^c|\vx_{1:T})$ is a Gaussian distribution with parameters mean $\bd{\mu}_c$ and diagonal variance matrix $\bd{\sigma}_c$. Inference model $q(\vz^m_t|\vx_{t}, \vz_{<t}^m)$ is a Gaussian distribution with parameters mean $\bd{\mu}_m$ and diagonal variance matrix $\bd{\sigma}_m$. For the prior distribution, we assume $p(\vz_{t}^m|\vz_{<t}^m)$ is a Gaussian distribution with parameters mean $\bd{\mu}_m^{\psi}$ and diagonal covariance matrix $\bd{\sigma}_m^{\psi}$.   
For regularizing the motion variables, we just use $\MMD$ without introducing any additional parameter, $\MMD_k(q(\vz_{t}^m|\vz_{<t}^m), p(\vz_{t}^m|\vz_{<t}^m))$, 
and we choose mixture of RBF kernel~\citep{li2017mmd}, where RBF kernel is defined as $k(\vx, \vy)=\exp(-\frac{\|\vx - \vy\|^2}{2\sigma^2}).$
With samples $\{\wt{\vz}_i\}_{i=1}^n$ from the posterior $q(\wt{\vz}^c)$ and samples $\{\vz_i\}_{i=1}^n$ from the prior $p(\vz^c),$ $\MMD_k(q(\wt{\vz}^c), p(\vz^c))$ is defined as
\begin{equation}
\label{def:mmd}
\MMD_k(q(\wt{\vz}^c), p(\vz^c))\!=\!\frac{1}{n(n-1)}\!\sum_{i\neq j} k(\vz_i, \vz_j)+\frac{1}{n(n-1)}\!\sum_{i\neq j} k(\wt{\vz}_i, \wt{\vz}_j)\!-\!\frac{1}{n^2}\!\sum_{i, j} k(\wt{\vz}_i, \vz_j).  
\end{equation}

The difference between R-WAE(MMD) and R-WAE(GAN) is how to choose metrics for the regularizer $\bbd(Q_{Z^c},\! P_{Z^c}),$ where $P_{Z^c}$ is the prior distribution and $Q_{Z^c}$ is the posterior distribution of the inference model.

\paragraph{R-WAE(MMD)} The regularizer $\bbd(Q_{Z^c},\! P_{Z^c})$ is chosen as, 
$$\bbd(Q_{Z^c},\! P_{Z^c})=\MMD_{k_\gamma} (Q(Z^c),\! P(Z^c)),$$
where the scaled MMD $\MMD_{k_\gamma} (Q(Z^c),\! P(Z^c))$ is chosen as
$$\MMD_{k_\gamma} (Q_{Z^c},\! P_{Z^c})=\frac{\widehat{\MMD}_{k_\gamma}(Q_{Z^c}, P_{Z^c})}{1+10\E_{\hat{P}}[\|\nabla f_{\gamma}(\vz^c)\|_F^2]},$$
where the function $f_{\gamma}(\vz^c)$ is the kernel feature map and $\widehat{\MMD}_{k_\gamma}(Q_{Z^c}, P_{Z^c})$ is defined in the following. When we have samples $\{\wt{\vz}_i^c\}_{i=1}^n$ from $Q(Z^c)$ and samples $\{\vz_i^c\}_{i=1}^n$ from $P(Z^c),$
\begin{align}
\widehat{\MMD}_{k_\gamma}(Q(Z^c), P(Z^c))& =\frac{1}{n(n-1)}\!\sum_{i\neq j} k( f_{\gamma}(\vz_i^c), f_{\gamma}(\vz_j^c))+\frac{1}{n(n-1)}\!\sum_{i\neq j} k(f_{\gamma}(\wt{\vz}_i^c), f_{\gamma}(\wt{\vz}_j^c))\\ \notag
&-\!\frac{1}{n^2}\!\sum_{i, j} k(f_{\gamma}(\wt{\vz}_i^c), f_{\gamma}(\vz_j^c)), 
\end{align}
where the RBF kernel $k$ is defined on scalar variables, $k(x,y)=\exp(-\frac{\|x - y\|^2}{2}).$ To avoid the situation where the generator gets stuck on a local optimum, we apply spectral parametrization for the weight matrix~\citep{miyato2018spectral}. The feature map $f_{\gamma}$ is updated $L$ steps at each iteration. To overcome posterior collapse and inference lagging, we will update the inference model per iteration of updating the decoder model for $L$ steps during training~\citep{he2019lagging}. See Algorithm~\ref{alg:alg1} for details.

\paragraph{R-WAE(GAN)} For the regularizer $\bbd_{\JS}(Q_{Z^c},\! P_{Z^c})$, we introduce a discriminator $D_{\gamma}.$ The loss is as follows,
\begin{equation}
\label{eq:ganopt}
\mathcal{L} = \E_{\vz^c\sim p(\vz^c)}[\log D_{\gamma}(\vz^c)] + \E_{\wt{\vz}^c\sim q(\wt{\vz}^c)}[\log(1-D_{\gamma}(\wt{\vz}^c)))], 
\end{equation}
where $p(\vz^c)$ is the prior distribution and $q(\wt{\vz}^c)$ is the posterior distribution of the inference model. To stabilize the training of the min-max problem in GAN-based optimization~\eqref{eq:ganopt}, a lot of stabilization techniques have been proposed~\citep{thanh2019improving, mescheder2018training, gulrajani2017improved, petzka2017regularization, roth2017stabilizing, qi2017loss}. Let samples $\{\vz^c\}$ are from the prior $p(\vz^c)$ and $\{\wt{\vz}^c\}$ are from the inference posterior $q(\wt{\vz}^c).$ In our R-WAE(GAN), we will adopt the regularization from \cite{ mescheder2018training} and \cite{thanh2019improving},
\begin{equation}
\mathcal{L}-\lambda \E[\|(\nabla D_{\gamma})_{\hat{\vz}^c}\|^2],    
\end{equation}
where $\hat{\vz}^c=\alpha\vz^c + (1-\alpha)\wt{\vz}^c$, $\alpha\in \mathcal{U}(0, 1)$ and $(\nabla D_{\gamma})_{\hat{\vz}^c}$ is evaluated its gradient at the point $\hat{\vz}^c$.

\begin{minipage}{0.48\textwidth}
\centering
\vspace{-0.3cm}
\begin{algorithm}[H] 
\caption{R-WAE(GAN) \label{alg:alg1}}  
\begin{algorithmic}
\STATE {\bf Input:} regularization coefficient $\beta$ and content prior $p(z^c)$
\STATE {\bf Goal}:~learn encoders $q_\ff(\vz^c|\vx_{1:T})$ and $q_\ff(\vz^m_t|\vx_t, \vz^m_{<t})$, prior $p_{\psi}(\vz^m_t|\vz^m_{<t})$, discriminator $D_\gamma$, and decoder $p_\theta(\vx_t|\vz_t),$ where $\vz_t=(\vz^c, \vz^m_t)$
\WHILE{not converged}
    \FOR{step 1 to L}
        \STATE Sample batch $X=\{\vx_t\}$
        \STATE Sample $\{\vz^c\}$ from prior $p(\vz^c)$ and $\{\vz_t^m\}$ from prior $p_{\psi}$
        \STATE Sample $\{\tilde{\vz}^c, \tilde{\vz}_t^m\}$ from encoders $q_\ff$
        \STATE Update discriminator $D_{\gamma}\!$ and encoders $q_\ff$ with loss given by \eqref{obj:loss},~\eqref{pen:gan}
    \ENDFOR
    % \STATE Sample batch $X=\{\vx_t\}$
    % \STATE Sample $\{\vz^c\}$ from prior $p(\vz^c)$ and $\{\vz_t^m\}$ from prior $p_{\psi}$
    % \STATE Sample $\{\tilde{\vz}^c, \tilde{\vz}_t^m\}$ from encoders $q_\ff$
    % \STATE Update discriminator $D_{\gamma}\!$ and encoders $q_\ff$ for $L$ steps with loss given by \eqref{obj:loss},~\eqref{pen:gan}
    \STATE Update $p_{\theta}$ and prior $p_\psi$ with loss given by \eqref{obj:loss} and \eqref{pen:gan}.
\ENDWHILE
\vspace{-.2\baselineskip}
\end{algorithmic}
\end{algorithm}
\vspace{-0.3cm}
\end{minipage}
\hfill
\begin{minipage}{0.48\textwidth}
\vspace{-0.3cm}
\begin{algorithm}[H]
\caption{R-WAE(MMD) \label{alg:alg2}}  
\begin{algorithmic}
\STATE {\bf Input:} regularization coefficient $\beta$ and content prior $p(z^c)$
\STATE {\bf Goal}: learn encoders $q_\ff(\vz^c|\vx_{1:T})$ and $q_\ff(\vz^m_t|\vx_t, \vz^m_{<t})$, prior $p_{\psi}(\vz^m_t|\vz^m_{<t})$, feature map $f_\gamma$ and decoder $p_{\theta}(\vx_t|\vz_t),$ where $\vz_t=(\vz^c, \vz^m_t)$
\WHILE{not converged}
    \FOR{step 1 to L}
        \STATE Sample batch $X=\{\vx_t\}$
        \STATE Sample $\{\vz^c\}$ from prior $p(\vz^c)$ and $\{\vz_t^m\}$ from prior $p_{\psi}$
        \STATE Sample $\{\tilde{\vz}^c, \tilde{\vz}_t^m\}$ from encoders $q_\ff$
        \STATE Update feature map $f_{\gamma}$ and encoders $q_\phi$ with loss given by \eqref{obj:loss},~\eqref{pen:mmd}
    \ENDFOR
    \STATE Update $p_{\theta}$ and prior $p_\psi$ with loss given by \eqref{obj:loss} and \eqref{pen:mmd}.
\ENDWHILE
\vspace{-.2\baselineskip}
\end{algorithmic}
\end{algorithm}
\vspace{-0.3cm}
\end{minipage}

\subsection{Appendix E: Unconditional Video Generation}

 \begin{figure*}[ht]
 \centering
 \begin{subfigure}{.3333\textwidth}
 \centering
 \begin{tabular}{ccccc}
 \animategraphics[loop,autoplay,height=.185\textwidth]{2}{figures/0G1fram}{0}{19}&\hspace{-0.5cm}
 \animategraphics[loop,autoplay,height=.185\textwidth]{2}{figures/2G0fram}{0}{19}&\hspace{-0.5cm}
 \animategraphics[loop,autoplay,height=.185\textwidth]{2}{figures/2G1fram}{0}{19}&\hspace{-0.5cm}
 \animategraphics[loop,autoplay,height=.185\textwidth]{2}{figures/3G0fram}{0}{19}&\hspace{-0.5cm}
 \animategraphics[loop,autoplay,height=.185\textwidth]{2}{figures/13G0fram}{0}{19} \\ 
 \animategraphics[loop,autoplay,height=.185\textwidth]{2}{figures/4G0fram}{0}{19}& \hspace{-0.5cm}
 \animategraphics[loop,autoplay,height=.185\textwidth]{2}{figures/11G0fram}{0}{19}&\hspace{-0.5cm}
 \animategraphics[loop,autoplay,height=.185\textwidth]{2}{figures/02G1fram}{0}{19}&\hspace{-0.5cm}
 \animategraphics[loop,autoplay,height=.185\textwidth]{2}{figures/05G0fram}{0}{19}&\hspace{-0.5cm}
 \animategraphics[loop,autoplay,height=.185\textwidth]{2}{figures/6G0fram}{0}{19}\\ 
 \animategraphics[loop,autoplay,height=.185\textwidth]{2}{figures/6G1fram}{0}{19}&\hspace{-0.5cm}
 \animategraphics[loop,autoplay,height=.185\textwidth]{2}{figures/04G0fram}{0}{19}&\hspace{-0.5cm} 
 \animategraphics[loop,autoplay,height=.185\textwidth]{2}{figures/7G0fram}{0}{19}&\hspace{-0.5cm}
 \animategraphics[loop,autoplay,height=.185\textwidth]{2}{figures/7G1fram}{0}{19}&\hspace{-0.5cm}
 \animategraphics[loop,autoplay,height=.185\textwidth]{2}{figures/07G0fram}{0}{19} \\
 \animategraphics[loop,autoplay,height=.185\textwidth]{2}{figures/8G0fram}{0}{19}&\hspace{-0.5cm}
 \animategraphics[loop,autoplay,height=.185\textwidth]{2}{figures/08G0fram}{0}{19}&\hspace{-0.5cm}
 \animategraphics[loop,autoplay,height=.185\textwidth]{2}{figures/8G1fram}{0}{19}&\hspace{-0.5cm}  
 \animategraphics[loop,autoplay,height=.185\textwidth]{2}{figures/10G1fram}{0}{19}&\hspace{-0.5cm}
 \animategraphics[loop,autoplay,height=.185\textwidth]{2}{figures/5G0fram}{0}{19}\\ 
 \animategraphics[loop,autoplay,height=.185\textwidth]{2}{figures/15G1fram}{0}{19}&\hspace{-0.5cm}
 \animategraphics[loop,autoplay,height=.185\textwidth]{2}{figures/14G0fram}{0}{19}&\hspace{-0.5cm}
 \animategraphics[loop,autoplay,height=.185\textwidth]{2}{figures/14G1fram}{0}{19}&\hspace{-0.5cm}
 \animategraphics[loop,autoplay,height=.185\textwidth]{2}{figures/12G0fram}{0}{19}& \hspace{-0.5cm}
 \animategraphics[loop,autoplay,height=.185\textwidth]{2}{figures/02G0fram}{0}{19}\\
 \end{tabular}
 \vspace{-0.2cm}
 \caption{R-WAE(MMD)}
 \end{subfigure}
 \hspace{-.3cm}
   \begin{subfigure}{.3333\textwidth}
   \centering
   \begin{tabular}{ccccc}
   \animategraphics[loop,autoplay,height=.185\textwidth]{2}{VAE_MNIST/01G0fram}{0}{19}&\hspace{-0.5cm}
  \animategraphics[loop,autoplay,height=.185\textwidth]{2}{VAE_MNIST/1G0fram}{0}{19}&\hspace{-0.5cm}
  \animategraphics[loop,autoplay,height=.185\textwidth]{2}{VAE_MNIST/1G1fram}{0}{19}&\hspace{-0.5cm}
  \animategraphics[loop,autoplay,height=.185\textwidth]{2}{VAE_MNIST/2G1fram}{0}{19}&\hspace{-0.5cm}
  \animategraphics[loop,autoplay,height=.185\textwidth]{2}{VAE_MNIST/18G0fram}{0}{19}\\
  \animategraphics[loop,autoplay,height=.185\textwidth]{2}{VAE_MNIST/19G0fram}{0}{19}& \hspace{-0.5cm}
  \animategraphics[loop,autoplay,height=.185\textwidth]{2}{VAE_MNIST/3G1fram}{0}{19}&\hspace{-0.5cm}
  \animategraphics[loop,autoplay,height=.185\textwidth]{2}{VAE_MNIST/03G0fram}{0}{19}&\hspace{-0.5cm}
  \animategraphics[loop,autoplay,height=.185\textwidth]{2}{VAE_MNIST/26G1fram}{0}{19}&\hspace{-0.5cm} 
  \animategraphics[loop,autoplay,height=.185\textwidth]{2}{VAE_MNIST/5G1fram}{0}{19}\\
  \animategraphics[loop,autoplay,height=.185\textwidth]{2}{VAE_MNIST/7G0fram}{0}{19}&\hspace{-0.5cm}
  \animategraphics[loop,autoplay,height=.185\textwidth]{2}{VAE_MNIST/7G1fram}{0}{19}& \hspace{-0.5cm} 
  \animategraphics[loop,autoplay,height=.185\textwidth]{2}{VAE_MNIST/8G1fram}{0}{19}&\hspace{-0.5cm}
  \animategraphics[loop,autoplay,height=.185\textwidth]{2}{VAE_MNIST/11G1fram}{0}{19}&\hspace{-0.5cm}
  \animategraphics[loop,autoplay,height=.185\textwidth]{2}{VAE_MNIST/2G0fram}{0}{19}\\
  \animategraphics[loop,autoplay,height=.185\textwidth]{2}{VAE_MNIST/12G1fram}{0}{19}&\hspace{-0.5cm}
  \animategraphics[loop,autoplay,height=.185\textwidth]{2}{VAE_MNIST/13G0fram}{0}{19}&\hspace{-0.5cm}
  \animategraphics[loop,autoplay,height=.185\textwidth]{2}{VAE_MNIST/13G1fram}{0}{19}& \hspace{-0.5cm} 
  \animategraphics[loop,autoplay,height=.185\textwidth]{2}{VAE_MNIST/15G0fram}{0}{19}&\hspace{-0.5cm}
  \animategraphics[loop,autoplay,height=.185\textwidth]{2}{VAE_MNIST/17G1fram}{0}{19}\\
  \animategraphics[loop,autoplay,height=.185\textwidth]{2}{VAE_MNIST/16G0fram}{0}{19}&\hspace{-0.5cm}
  \animategraphics[loop,autoplay,height=.185\textwidth]{2}{VAE_MNIST/16G1fram}{0}{19}&\hspace{-0.5cm}
  \animategraphics[loop,autoplay,height=.185\textwidth]{2}{VAE_MNIST/19G1fram}{0}{19}& \hspace{-0.5cm}
  \animategraphics[loop,autoplay,height=.185\textwidth]{2}{VAE_MNIST/12G0fram}{0}{19}&\hspace{-0.5cm}
  \animategraphics[loop,autoplay,height=.185\textwidth]{2}{VAE_MNIST/3G0fram}{0}{19}\\
  \end{tabular}
   \vspace{-0.2cm}
  \caption{DS-VAE} 
  \end{subfigure}
 \hspace{-.3cm}
  \begin{subfigure}{.3333\textwidth}
  \centering
  \begin{tabular}{ccccc}
  \animategraphics[loop,autoplay,height=.185\textwidth]{2}{mocogan_smmnist/0_smmnist_}{0}{19}&\hspace{-0.5cm}
  \animategraphics[loop,autoplay,height=.185\textwidth]{2}{mocogan_smmnist/1_smmnist_}{0}{19}&\hspace{-0.5cm}
  \animategraphics[loop,autoplay,height=.185\textwidth]{2}{mocogan_smmnist/2_smmnist_}{0}{19}&\hspace{-0.5cm}
  \animategraphics[loop,autoplay,height=.185\textwidth]{2}{mocogan_smmnist/3_smmnist_}{0}{19}&\hspace{-0.5cm}
  \animategraphics[loop,autoplay,height=.185\textwidth]{2}{mocogan_smmnist/4_smmnist_}{0}{19}\\ 
  \animategraphics[loop,autoplay,height=.185\textwidth]{2}{mocogan_smmnist/5_smmnist_}{0}{19}&\hspace{-0.5cm}
  \animategraphics[loop,autoplay,height=.185\textwidth]{2}{mocogan_smmnist/6_smmnist_}{0}{19}&\hspace{-0.5cm}
  \animategraphics[loop,autoplay,height=.185\textwidth]{2}{mocogan_smmnist/7_smmnist_}{0}{19}&\hspace{-0.5cm} 
  \animategraphics[loop,autoplay,height=.185\textwidth]{2}{mocogan_smmnist/8_smmnist_}{0}{19}&\hspace{-0.5cm}
  \animategraphics[loop,autoplay,height=.185\textwidth]{2}{mocogan_smmnist/9_smmnist_}{0}{19} \\
  \animategraphics[loop,autoplay,height=.185\textwidth]{2}{mocogan_smmnist/10_smmnist_}{0}{19}&\hspace{-0.5cm}
  \animategraphics[loop,autoplay,height=.185\textwidth]{2}{mocogan_smmnist/11_smmnist_}{0}{19}&\hspace{-0.5cm}
  \animategraphics[loop,autoplay,height=.185\textwidth]{2}{mocogan_smmnist/12_smmnist_}{0}{19}&\hspace{-0.5cm}
  \animategraphics[loop,autoplay,height=.185\textwidth]{2}{mocogan_smmnist/13_smmnist_}{0}{19}&\hspace{-0.5cm}
  \animategraphics[loop,autoplay,height=.185\textwidth]{2}{mocogan_smmnist/14_smmnist_}{0}{19} \\
  \animategraphics[loop,autoplay,height=.185\textwidth]{2}{mocogan_smmnist/15_smmnist_}{0}{19}&\hspace{-0.5cm}
  \animategraphics[loop,autoplay,height=.185\textwidth]{2}{mocogan_smmnist/16_smmnist_}{0}{19}&\hspace{-0.5cm}
  \animategraphics[loop,autoplay,height=.185\textwidth]{2}{mocogan_smmnist/17_smmnist_}{0}{19}&\hspace{-0.5cm}
  \animategraphics[loop,autoplay,height=.185\textwidth]{2}{mocogan_smmnist/18_smmnist_}{0}{19}&\hspace{-0.5cm}
  \animategraphics[loop,autoplay,height=.185\textwidth]{2}{mocogan_smmnist/19_smmnist_}{0}{19} \\
  \animategraphics[loop,autoplay,height=.185\textwidth]{2}{mocogan_smmnist/20_smmnist_}{0}{19}&\hspace{-0.5cm}
  \animategraphics[loop,autoplay,height=.185\textwidth]{2}{mocogan_smmnist/21_smmnist_}{0}{19}&\hspace{-0.5cm}
  \animategraphics[loop,autoplay,height=.185\textwidth]{2}{mocogan_smmnist/22_smmnist_}{0}{19}&\hspace{-0.5cm}
  \animategraphics[loop,autoplay,height=.185\textwidth]{2}{mocogan_smmnist/23_smmnist_}{0}{19}&\hspace{-0.5cm}
  \animategraphics[loop,autoplay,height=.185\textwidth]{2}{mocogan_smmnist/24_smmnist_}{0}{19} \\ 
  \end{tabular}
  \vspace{-0.2cm}
  \caption{MoGoGAN} 
  \end{subfigure}

  \caption{Unconditional video generation on SM-MNIST: (a) Sequences (length=20) in R-WAE(MMD) are randomly taken from generated samples with $T=100$ to save pdf size; (b) Generated videos by DS-VAE~\citep{li2018disentangled} with $T=20$; (c) Generated videos by MoCoGAN~ \citep{tulyakov2018mocogan} with $T=20$. The figures should be viewed with Adobe Reader to see video. \label{fig:mnistsample}
 %\caption{Unconditional video generation on SM-MNIST. (a), sequences (length=20) in R-WAE(MMD) are randomly taken from generated samples with $T=100$ to save pdf size. \label{fig:mnistsample}
 %\textbf{Erran: (b) and (c) are missing}
 } 
 \end{figure*}

Fig.~\ref{fig:mnistsample} provides generated samples on the SM-MNIST dataset by randomly sampling content $\{\vz^c\}$ from the prior $p(\vz^c)$ and motions $\{\vz^m_{1:T}\}$ from the learned prior $p_\psi(\vz_t^m|\vz_{<t}^m).$ The length of our generated videos is $T=100$ and we only show randomly chosen videos of $T=20$ to save file size. Our R-WAE(MMD) achieves the most consistent and visually best sequence even when $T=100$. Samples from MoCoGAN~\citep{tulyakov2018mocogan} usually change digit identity along the sequence. The reason is that MoCoGAN~\citep{tulyakov2018mocogan} requires the number of actions be finite.

Fig.~\ref{fig:mugsample} shows unconditional video generation with $T=10$ on MUG facial dataset. DS-VAE in (b) is improved by incorporating categorical latent variables. The figures should be viewed with Adobe Reader to see video.
\begin{figure*}[ht]
\centering
\begin{subfigure}{0.33333\textwidth}
\centering
\begingroup
\begin{tabular}{ccccc}
\animategraphics[loop,autoplay,height=.185\textwidth]{1}{figures/R0G1fram}{0}{9}&\hspace{-0.5cm}
\animategraphics[loop,autoplay,height=.185\textwidth]{1}{figures/R01G1fram}{0}{9}&\hspace{-0.5cm}
\animategraphics[loop,autoplay,height=.185\textwidth]{1}{figures/R2G0fram}{0}{9}&\hspace{-0.5cm}
\animategraphics[loop,autoplay,height=.185\textwidth]{1}{figures/R02G1fram}{0}{9}&\hspace{-0.5cm}
\animategraphics[loop,autoplay,height=.185\textwidth]{1}{figures/R6G1fram}{0}{9}\\
\animategraphics[loop,autoplay,height=.185\textwidth]{1}{figures/R7G0fram}{0}{9}&\hspace{-0.5cm}
\animategraphics[loop,autoplay,height=.185\textwidth]{1}{figures/R8G0fram}{0}{9}&\hspace{-0.5cm}
\animategraphics[loop,autoplay, height=.185\textwidth]{1}{figures/R8G1fram}{0}{9}&\hspace{-0.5cm}
\animategraphics[loop,autoplay,height=.185\textwidth]{1}{figures/R9G0fram}{0}{9}&\hspace{-0.5cm}
\animategraphics[loop,autoplay,height=.185\textwidth]{1}{figures/R11G1fram}{0}{9}\\ 
\animategraphics[loop,autoplay,height=.185\textwidth]{1}{figures/R12G1fram}{0}{9}&\hspace{-0.5cm}
\animategraphics[loop,autoplay,height=.185\textwidth]{1}{figures/R13G0fram}{0}{9}&\hspace{-0.5cm}
\animategraphics[loop,autoplay,height=.185\textwidth]{1}{figures/R14G0fram}{0}{9}&\hspace{-0.5cm}
\animategraphics[loop,autoplay,height=.185\textwidth]{1}{figures/R16G0fram}{0}{9}&\hspace{-0.5cm}
\animategraphics[loop,autoplay,height=.185\textwidth]{1}{figures/R19G0fram}{0}{9}\\ 
\animategraphics[loop,autoplay,height=.185\textwidth]{1}{figures/R15G1fram}{0}{9}&\hspace{-0.5cm}
\animategraphics[loop,autoplay,height=.185\textwidth]{1}{figures/R14G1fram}{0}{9}&\hspace{-0.5cm}
\animategraphics[loop,autoplay,height=.185\textwidth]{1}{figures/R3G1fram}{0}{9}&\hspace{-0.5cm}
\animategraphics[loop,autoplay,height=.185\textwidth]{1}{figures/R6G0fram}{0}{9}&\hspace{-0.5cm}
\animategraphics[loop,autoplay,height=.185\textwidth]{2}{figures/R13G1fram}{0}{9}\\ 
\animategraphics[loop,autoplay,height=.185\textwidth]{1}{figures/R11G0fram}{0}{9}&\hspace{-0.5cm}
\animategraphics[loop,autoplay,height=.185\textwidth]{1}{figures/R7G1fram}{0}{9}&\hspace{-0.5cm}
\animategraphics[loop,autoplay,height=.185\textwidth]{1}{figures/R5G1fram}{0}{9}&\hspace{-0.5cm}
\animategraphics[loop,autoplay,height=.185\textwidth]{1}{figures/R23G1fram}{0}{9}&\hspace{-0.5cm}
\animategraphics[loop,autoplay,height=.185\textwidth]{1}{figures/R10G0fram}{0}{9} \\
\end{tabular}
\endgroup
\vspace{-0.2cm}
\caption{R-WAE(GAN)}
\end{subfigure}%
\hspace{-.3cm}
\begin{subfigure}{0.33333\textwidth}
\centering
\begin{tabular}{ccccc}
\animategraphics[loop,autoplay,height=.185\textwidth]{1}{figures/VAE/0G0fram}{0}{9}&\hspace{-0.5cm}
\animategraphics[loop,autoplay,height=.185\textwidth]{1}{figures/VAE/0G1fram}{0}{9}&\hspace{-0.5cm}
\animategraphics[loop,autoplay,height=.185\textwidth]{1}{figures/VAE/3G1fram}{0}{9}&\hspace{-0.5cm}
\animategraphics[loop,autoplay,height=.185\textwidth]{1}{figures/VAE/4G1fram}{0}{9}&\hspace{-0.5cm}
\animategraphics[loop,autoplay,height=.185\textwidth]{1}{figures/VAE/5G1fram}{0}{9}\\ 
\animategraphics[loop,autoplay,height=.185\textwidth]{1}{figures/VAE/7G0fram}{0}{9}&\hspace{-0.5cm}
\animategraphics[loop,autoplay,height=.185\textwidth]{1}{figures/VAE/7G1fram}{0}{9}&\hspace{-0.5cm}
\animategraphics[loop,autoplay, height=.185\textwidth]{1}{figures/VAE/8G0fram}{0}{9}&\hspace{-0.5cm}
\animategraphics[loop,autoplay,height=.185\textwidth]{1}{figures/VAE/9G1fram}{0}{9}&\hspace{-0.5cm}
\animategraphics[loop,autoplay,height=.185\textwidth]{1}{figures/VAE/10G0frame}{0}{9}\\
\animategraphics[loop,autoplay,height=.185\textwidth]{1}{figures/VAE/10G1fram}{0}{9}&\hspace{-0.5cm}
\animategraphics[loop,autoplay,height=.185\textwidth]{1}{figures/VAE/11G0frame}{0}{9}&\hspace{-0.5cm}
\animategraphics[loop,autoplay,height=.185\textwidth]{2}{figures/VAE/19G0frame}{0}{9}&\hspace{-0.5cm}
\animategraphics[loop,autoplay,height=.185\textwidth]{1}{figures/VAE/13G0frame}{0}{9}&\hspace{-0.5cm}
\animategraphics[loop,autoplay,height=.185\textwidth]{1}{figures/VAE/15G1frame}{0}{9}\\ 
\animategraphics[loop,autoplay,height=.185\textwidth]{1}{figures/VAE/16G0fram}{0}{9}&\hspace{-0.5cm}
\animategraphics[loop,autoplay,height=.185\textwidth]{1}{figures/VAE/17G0fram}{0}{9}&\hspace{-0.5cm}
\animategraphics[loop,autoplay,height=.185\textwidth]{1}{figures/VAE/18G0fram}{0}{9}&\hspace{-0.5cm}
\animategraphics[loop,autoplay,height=.185\textwidth]{2}{figures/VAE/14G0fram}{0}{9}&\hspace{-0.5cm}
\animategraphics[loop,autoplay,height=.185\textwidth]{1}{figures/VAE/11G1fram}{0}{9}\\ 
\animategraphics[loop,autoplay,height=.185\textwidth]{1}{figures/VAE/19G1frame}{0}{9}&\hspace{-0.5cm}
\animategraphics[loop,autoplay,height=.185\textwidth]{1}{figures/VAE/20G0frame}{0}{9}&\hspace{-0.5cm}
\animategraphics[loop,autoplay,height=.185\textwidth]{1}{figures/VAE/21G0frame}{0}{9}&\hspace{-0.5cm}
\animategraphics[loop,autoplay,height=.185\textwidth]{1}{figures/VAE/21G1frame}{0}{9}&\hspace{-0.5cm}
\animategraphics[loop,autoplay,height=.185\textwidth]{1}{figures/VAE/22G0frame}{0}{9} \\
\end{tabular}
\vspace{-0.2cm}
\caption{DS-VAE}
\end{subfigure}
\hspace{-.3cm}
\begin{subfigure}{0.33333\textwidth}
\centering
\label{fig:mug_mocogan}
\begingroup
\begin{tabular}{ccccc}
\animategraphics[loop,autoplay,height=.185\textwidth]{1}{mocogan_mug/0_mug_}{0}{9}&\hspace{-0.5cm}
\animategraphics[loop,autoplay,height=.185\textwidth]{1}{mocogan_mug/1_mug_}{0}{9}&\hspace{-0.5cm}
% \animategraphics[loop,autoplay,height=.18\textwidth]{1}{mocogan_mug/2_mug_}{0}{9}&\hspace{-0.5cm}
\animategraphics[loop,autoplay,height=.185\textwidth]{1}{mocogan_mug/3_mug_}{0}{9}&\hspace{-0.5cm}
\animategraphics[loop,autoplay,height=.185\textwidth]{1}{mocogan_mug/4_mug_}{0}{9}&\hspace{-0.5cm}
\animategraphics[loop,autoplay,height=.185\textwidth]{1}{mocogan_mug/5_mug_}{0}{9}\\ 
\animategraphics[loop,autoplay,height=.185\textwidth]{1}{mocogan_mug/6_mug_}{0}{9}&\hspace{-0.5cm}
\animategraphics[loop,autoplay,height=.185\textwidth]{1}{mocogan_mug/7_mug_}{0}{9}&\hspace{-0.5cm}
\animategraphics[loop,autoplay,height=.185\textwidth]{1}{mocogan_mug/8_mug_}{0}{9}&\hspace{-0.5cm}
\animategraphics[loop,autoplay,height=.185\textwidth]{1}{mocogan_mug/9_mug_}{0}{9}&\hspace{-0.5cm}
\animategraphics[loop,autoplay,height=.185\textwidth]{1}{mocogan_mug/10_mug_}{0}{9}\\ 
% \animategraphics[loop,autoplay,height=.12\textwidth]{1}{mocogan_mug/11_mug_}{0}{9}\\
\animategraphics[loop,autoplay,height=.185\textwidth]{1}{mocogan_mug/12_mug_}{0}{9}&\hspace{-0.5cm}
\animategraphics[loop,autoplay,height=.185\textwidth]{1}{mocogan_mug/13_mug_}{0}{9}&\hspace{-0.5cm}
\animategraphics[loop,autoplay,height=.185\textwidth]{1}{mocogan_mug/14_mug_}{0}{9}&\hspace{-0.5cm}
\animategraphics[loop,autoplay,height=.185\textwidth]{1}{mocogan_mug/15_mug_}{0}{9}&\hspace{-0.5cm}
% \animategraphics[loop,autoplay,height=.18\textwidth]{1}{mocogan_mug/16_mug_}{0}{9}&\hspace{-0.5cm}
\animategraphics[loop,autoplay,height=.185\textwidth]{1}{mocogan_mug/17_mug_}{0}{9}\\ 
\animategraphics[loop,autoplay,height=.185\textwidth]{1}{mocogan_mug/18_mug_}{0}{9}&\hspace{-0.5cm}
\animategraphics[loop,autoplay,height=.185\textwidth]{1}{mocogan_mug/19_mug_}{0}{9}&\hspace{-0.5cm}
\animategraphics[loop,autoplay,height=.185\textwidth]{1}{mocogan_mug/20_mug_}{0}{9}&\hspace{-0.5cm}
% \animategraphics[loop,autoplay,height=.12\textwidth]{1}{mocogan_mug/21_mug_}{0}{9}&\hspace{-0.5cm}
\animategraphics[loop,autoplay,height=.185\textwidth]{2}{mocogan_mug/22_mug_}{0}{9}&\hspace{-0.5cm}
\animategraphics[loop,autoplay,height=.185\textwidth]{2}{mocogan_mug/23_mug_}{0}{9}\\ 
\animategraphics[loop,autoplay,height=.185\textwidth]{1}{mocogan_mug/24_mug_}{0}{9}&\hspace{-0.5cm}
\animategraphics[loop,autoplay,height=.185\textwidth]{1}{mocogan_mug/25_mug_}{0}{9}&\hspace{-0.5cm}
\animategraphics[loop,autoplay,height=.185\textwidth]{1}{mocogan_mug/26_mug_}{0}{9}&\hspace{-0.5cm}
% \animategraphics[loop,autoplay,height=.12\textwidth]{1}{mocogan_mug/27_mug_}{0}{9}&\hspace{-0.5cm}
\animategraphics[loop,autoplay,height=.185\textwidth]{1}{mocogan_mug/28_mug_}{0}{9}&\hspace{-0.5cm}
\animategraphics[loop,autoplay,height=.185\textwidth]{1}{mocogan_mug/29_mug_}{0}{9} \\
\end{tabular}
\endgroup 
\vspace{-0.2cm}
\caption{MoCoGAN}
\end{subfigure}
\vspace{-0.2cm}
\caption{Unconditional video generation with $T=10$ on MUG facial dataset. DS-VAE in (b) is improved by incorporating categorical latent variables. The figures should be viewed with Adobe Reader to see video. \label{fig:mugsample}} 
\end{figure*}

\subsection*{Appendix F: Latent Manifold Visualization}
\begin{figure}[ht]
    \centering
   \includegraphics[height=0.6\textwidth]{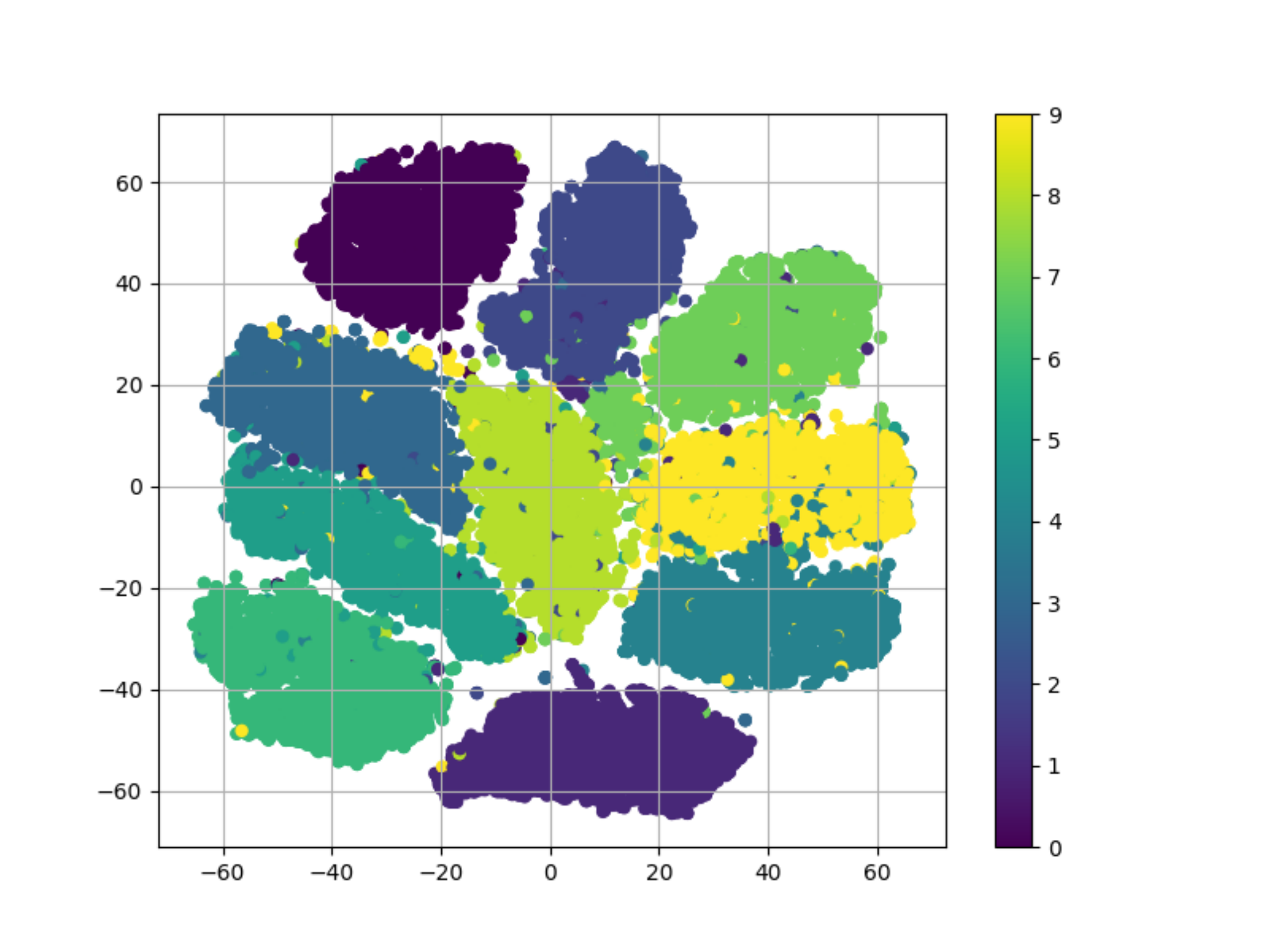}   
    \caption{Visualizing 2D manifold of content code $\{\vz^c\}$ encoded from R-WAE(MMD) on SM-MNIST by t-SNE~\citep{maaten2008visualizing}.
    \label{manifold}}
 \end{figure}

We encode the test data $\{\vx_{1:T}\}$ of SM-MNIST with $T=10$ to get the content codes $\{\vz^c\}$ using our R-WAE(MMD). We visualize two-dimensional (2D) manifold of $\{\vz^c\}$ using t-SNE~\citep{maaten2008visualizing}. In Fig.~\ref{manifold}, different colors correspond to the digit identities of the latent codes $\{\vz^c\}$ of test videos on SM-MNIST. This indicates that $\{\vz^c\}$ encoded by our R-WAE(MMD) exactly captures the invariant information (digits) of the test data. The latent motion codes are sequential and cannot be visualized.

\subsection*{Appendix G: Model Architecture and Hyper-parameters}
\begin{figure*}[ht]
\centering
\begin{tabular}{cc}
\includegraphics[height=0.5\textwidth]{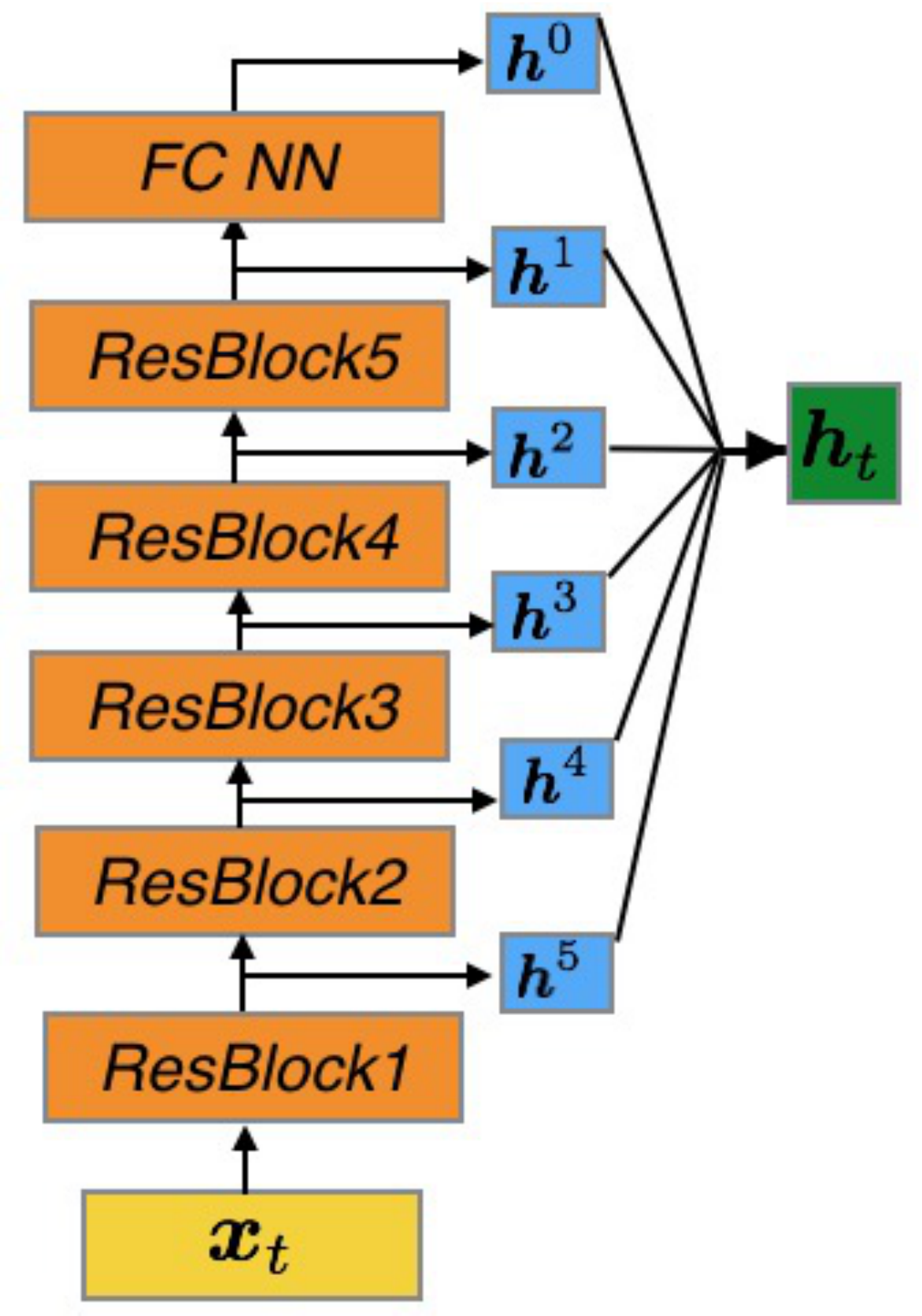}& 
\includegraphics[height=0.5\textwidth]{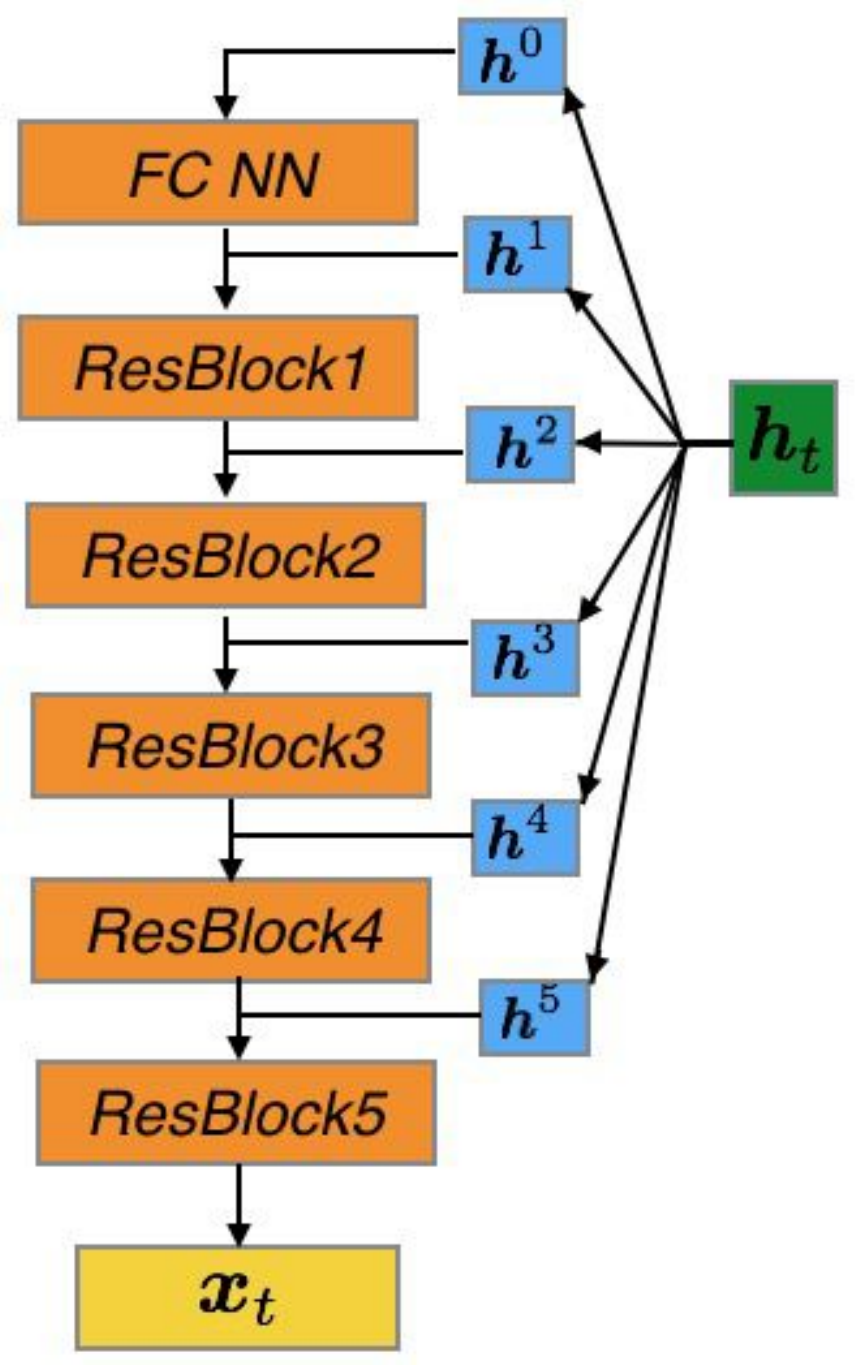} \\
{\small \!\!\!\! (a) Encoder Network} & {\small\!\!\!\! (b) Decoder Network}
\end{tabular}
\caption{Structures of the encoder network and decoder network. (a) The ResBlock in the encoder network consists of convolutional network adopted from~\cite{brock2018large}, named "ResBlock down". After each Resblock, we use a FC network to get latent feature $\vh^i$, for $i=0,\cdots, 5$ (Ladder Network~\citep{sonderby2016ladder, zhao2017learning}), whose dimensions are the same. $[\vh^5, \vh^4, \vh^3, \vh^2, \vh^1, \vh^0]$ are concatenated into latent feature $\vh_t$, where $\vh_t$ is defined in Fig.\ref{fig:struct}. We use deconvolutional network adopted from~\cite{brock2018large}, named "ResBlock up". In (b), the hidden state $\vh_t$ of an LSTM, defined in Fig.\ref{fig:struct},  is evenly split into $[\vh^5, \vh^4, \vh^3, \vh^2, \vh^1, \vh^0]$. And the ResBlock in decoder network consists of deconvolutional network adopted from~\cite{brock2018large}. We use leaky relu activation for all ResBlocks. \label{fig:encdec}}
\end{figure*}

\begin{figure*}[ht]
\centering
\begin{tabular}{ccccc}
\includegraphics[height=0.25\textwidth]{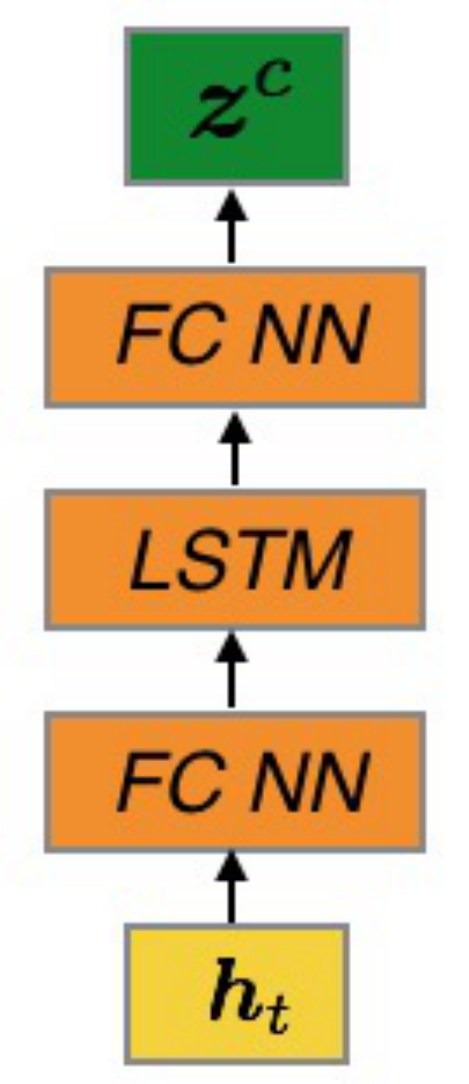}& 
\includegraphics[height=0.22\textwidth]{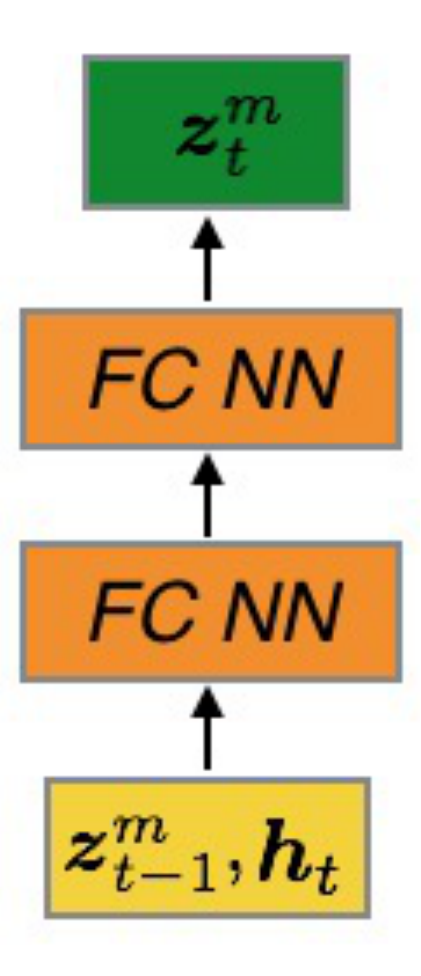}&
\includegraphics[height=0.25\textwidth]{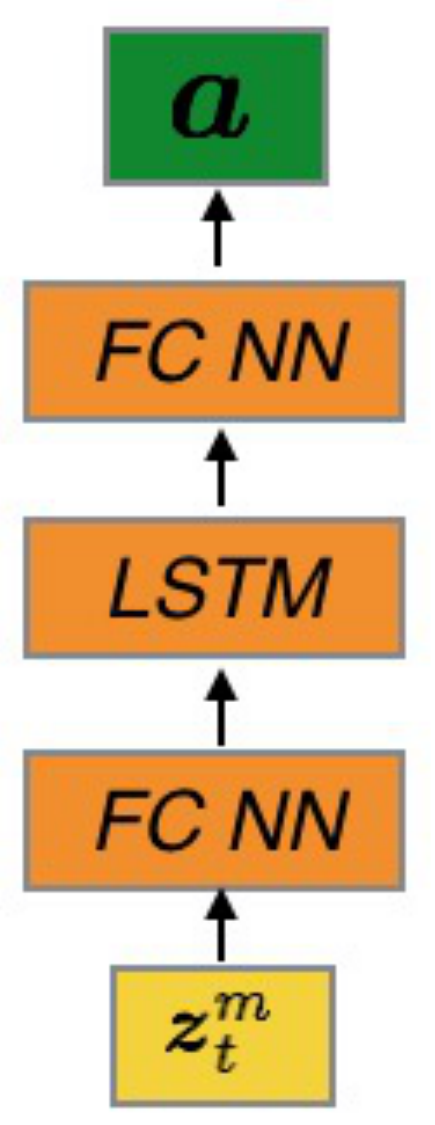}&
\includegraphics[height=0.23\textwidth]{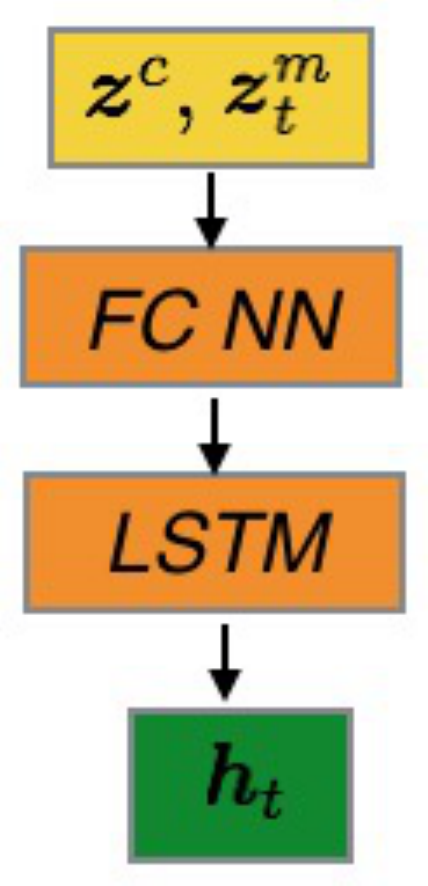}&
\includegraphics[height=0.23\textwidth]{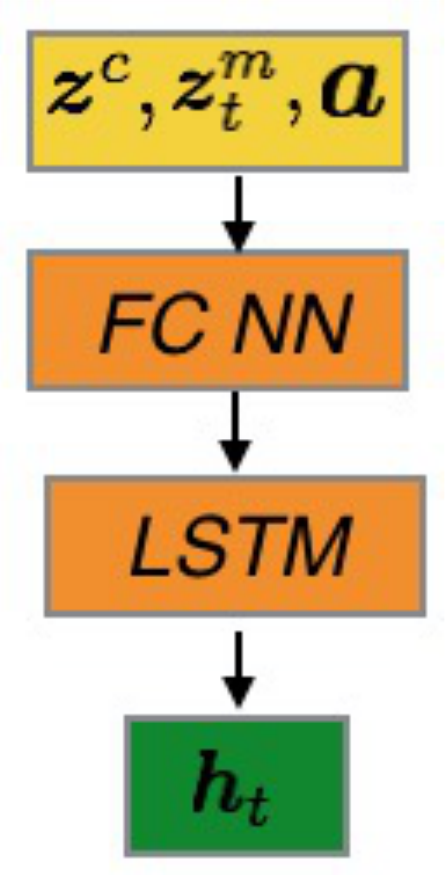}\\
{\small \!\!\!\! (a) infer $\vz^c$ } & {\small\!\!\!\! (b) infer $\vz^m_t$}&
{\small\!\!\!\! (c) infer $\va$}&{\small\!\!\!\! (d) output $\vh_t$ for decoder}&
{\small\!\!\!\! (e) output $\vh_t$ for weakly-supervised decoder}
\end{tabular}
\caption{Network architectures in addition to encoder/decoder network with $\vh_t$ defined in Fig.~\ref{fig:encdec}. (a) Network structure to infer the content variable $\vz^c$ from sequence $\vx_{1:T}$; (b) Network structure to infer content variable $\vz^m_t$; (c) In inference model, we introduce an additional Gumbel random variable $\va$ inferred by motion sequences $\{\vz^m_t\};$ (d) Content variable $\vz^c$ and motion variable $\vz^m_t$ are concatenated into an LSTM for the decoder model; (e) In weakly-supervised inference model, content variable $\vz^c$, motion variable $\vz^m_t$ and Gumbel random variable $\va$ are concatenated into an LSTM for the decoder model. \label{fig:contmot}}
\end{figure*}

In the inference model, we use an encoder network, defined in Fig.~\ref{fig:encdec} (a) to extract latent feature $\vh_t$ defined in Fig.\ref{fig:struct}. We use a decoder network to reconstruct $\hat{\vx}_t$ from the hidden state $\vh_t$, defined in Fig.\ref{fig:struct}. For the discriminator $D_{\gamma}$ in R-WAE(GAN), we use a 4-layer fully-connected neural network (FC NN) with respective dimension $(256, 256, 128, 1).$ For the feature map $f_{\gamma}$ with a scalar output for the RBF kernel of R-WAE(MMD), we use a 4-layer fully-connected neural network with respective dimension $(256, 256, 128, 1).$ After encoding $\vx_t$, we get extracted latent feature $\vh_t.$ We use Fig.~\ref{fig:contmot}(a) and Fig.~\ref{fig:contmot}(b) to infer the content variable $\vz^c$ and motion variables $\vz^m_t.$ When the Gumbel latent variable is incorporated into our weakly-supervised inference model, we use Fig.~\ref{fig:contmot}(c) to infer the Gumbel latent variable $\va$. The latent content variable $\vz^c$ and latent motion variable $\vz^m_t$ are concatenated as input to an LSTM after an FC NN to output hidden state $\vh_t$ for reconstructing $\hat{\vx}_t$ using the decoder. For our weakly-supervised model, the latent content variable $\vz^c,$ latent motion variable $\vz^m_t$ and latent action variable $\va$ are concatenated as input to an LSTM after an FC NN to output hidden state $\vh_t$ for reconstructing $\hat{\vx}_t$ using the decoder. We use Adam optimizer~\citep{kingma2014adam} with $\beta^1=0.5$ and $\beta^2=0.9.$

\begin{table*}[h!]
    %\vspace{-0.05in}
    \centering
    \scalebox{1}{
    % \resizebox{1\textwidth}{!}{
    \begin{tabular}{l|cc}
    %\hline
    \multicolumn{1}{l}{} & \multicolumn{2}{c}{{\bf Sprites}} \\
    \hline
    Methods  & actions &  content \\
    \hline
    R-WAE(GAN) (S) & 3.73\% & 2.00\% \\
    R-WAE(MMD) (S) & 5.83\% & 2.45\% \\
    R-WAE(GAN) (C) & 3.13\% & 3.31\% \\
    R-WAE(MMD) (C) & 7.72\% & 3.31\% \\
    \hline
    \end{tabular}}
    \caption{Results of R-WAE(GAN) and R-WAE(MMD) on Sprites dataset.}
    \label{tab:mmd_gan}
    %\vspace{-0.05in}
\end{table*}

\paragraph{Architecture on SM-MNIST, Sprites and TIMIT Datasets} We use the same architecture on SM-MNIST and Sprites dataset, as shown in Fig.~\ref{fig:paramtoy}. The details of the parameters of the networks are provided in Fig.~\ref{fig:paramtoy}. As R-WAE(GAN) and R-WAE(MMD) have similar performance on SM-MNIST and Sprites (see Sprites results in Table 4), we only provide the results and parameters of R-WAE(MMD) to save space. At each iteration of training the decoder $p_{\theta}(\vx_t|\vz_t)$ and the prior $p_{\psi}(\vz_t^m|\vz_{<t}^m)$, we train the encoder parameters $q_{\phi}$ and the feature map $f_{\gamma}$ for R-WAE(MMD) with $L$ steps. The results on SM-MNIST and Sprites datasets are evaluated after 500 epochs. On SM-MNIST dataset, we use a Bernoulli cross-entropy loss and choose $L=5$. The penalty coefficients $\beta_1$ and $\beta_2$, are, respectively, $5$ and $20$. The learning rate for the decoder model is $5\times 10^{-4}$ and the learning rate for the encoder is $1\times 10^{-4}.$ The learning rate for $f_{\gamma}$ is $1\times 10^{-4}.$  On Sprites dataset, we use an $L2$ reconstruction loss and choose $L=5$ steps. The penalty coefficients $\beta_1$ and $\beta_2$ are, respectively, $10$ and $60$.  The learning rate for the decoder model is $3\times 10^{-4}$ and the learning rate for the encoder is $1\times 10^{-4}.$ The learning rate for $D_{\gamma}$ in R-WAE(GAN) or $f_{\gamma}$ in R-WAE(MMD) is $1\times 10^{-4}.$ We use a decayed learning rate schedule on both datasets. After 50 epochs, we decrease all learning rates by a factor of 2 and after 80 epochs decrease further by a factor of 5. On TIMIT speech dataset, we use the same encoder and decoder architecture as that of DS-VAE. The dimension of hidden states is 256 and the dimensions of $\vz^c$ and $\vz_t^m$ are both 16. 

\begin{figure}
 \begin{minipage}[h]{0.48\textwidth}
    \centering
     \begin{tabular}{c}
    \hline\hline
    ResBlock1 down 64*3*3 \\ \hline
    self-attention \\ \hline
    ResBlock2 down 128*3*3 \\ \hline
    ResBlock3 down 256*3*3 \\ \hline
    ResBlock4 down 512*3*3 \\ \hline
    ResBlock5 down 1024*3*3 \\ \hline
    Reshape output to $(N, 1024\times2\times2)$\\\hline 
    FC NN \\ \hline\hline \\
    \end{tabular}
  \captionof{table}{Encoder Network Architecture.}    
  \end{minipage}
  \hfill \hspace{-0.1cm}
  \begin{minipage}[bth]{0.48\textwidth}
  \begin{tabular}{c}
    \hline\hline
    FC NN and Reshape input to (N, 2048, 2, 2) \\\hline
    ResBlock1 up 1024*3*3 \\ \hline
    ResBlock2 up 512*3*3 \\ \hline
    ResBlock3 up 256*3*3 \\ \hline
    ResBlock4 up 128*3*3 \\ \hline
    self-attention \\ \hline
    ResBlock5 up 64*3*3 \\ \hline
    Conv 3*3*3, activation=sigmoid
     \\ \hline\hline \\
\end{tabular}
\captionof{table}{Decoder Network Architecture.}
\end{minipage}
\vspace{-0.1cm}
\caption{Network parameters on encoder network and decoder network on SM-MNIST and Sprites datasets. We adopt ResBlock down and up from \cite{brock2018large}. The dimensions of $\vz^c$, $\vz^m_t$, $\vh_t$ are $120$, $12$ and $150$ respectively. The batch size on both SM-MNIST and Sprites dataset are $60$ and the length of video sequence for training is $T=8.$\label{fig:paramtoy}} 
\end{figure}

\paragraph{Architecture on MUG Facial Dataset} The details of the architecture parameters of the networks for MUG facial dataset are provided in Fig.~\ref{fig:paramtoy}.  The results on MUG facial dataset are evaluated after 800 epochs. For the regularizer $\bbd_{\KL}(q_{\phi}(\va|\vx_{1:T},\vz^m_{1:T}), p(\va))$, we choose the coefficient of this categorical regularizer to be $50$. We use an $L2$ reconstruction loss and choose $L=5$ steps. For R-WAE(MMD), the penalty coefficients $\beta_1$ and $\beta_2$ are, respectively, $10$ and $50$. For R-WAE(GAN), the coefficients $\beta_1$ and $\beta_2$ of the penalties are, respectively, $5$ and $60$. The learning rate for the decoder model is $5\times 10^{-4}$ and the learning rate for the encoder is $2\times 10^{-4}.$ The learning rate for $D_{\gamma}$ in R-WAE(GAN) or $f_{\gamma}$ in R-WAE(MMD) is $2\times 10^{-4}.$ We use the same decayed learning rate schedule as described on SM-MNIST and Sprites datasets. This architecture can be applied to improve the compression rate~\citep{han2018deep}.

\begin{figure}
 \begin{minipage}[h]{0.48\textwidth}
    \centering
     \begin{tabular}{c}
    \hline\hline
    ResBlock1 down 64*3*3 \\ \hline
    self-attention \\ \hline
    ResBlock2 down 128*3*3 \\ \hline
    ResBlock3 down 256*3*3 \\ \hline
    ResBlock4 down 512*3*3 \\ \hline
    ResBlock5 down 1024*3*3 \\ \hline
    Reshape output to $(N, 1024\times2\times2)$\\\hline 
    FC NN \\ \hline\hline \\
    \end{tabular}
  \captionof{table}{Encoder Network Architecture.}    
  \end{minipage}
  \hfill \hspace{-0.1cm}
  \begin{minipage}[bth]{0.48\textwidth}
  \begin{tabular}{c}
    \hline\hline
    FC NN and Reshape to (N, 3072, 2, 2) \\\hline
    ResBlock1 up 1536*3*3 \\ \hline
    ResBlock2 up 768*3*3 \\ \hline
    ResBlock3 up 384*3*3 \\ \hline
    ResBlock4 up 192*3*3 \\ \hline
    self-attention \\ \hline
    ResBlock5 up 96*3*3 \\ \hline
    Conv 3*3*3, activation=sigmoid
     \\ \hline\hline \\
\end{tabular}
\captionof{table}{Decoder Network Architecture.}
\end{minipage}
\vspace{-0.1cm}
\caption{Network parameters on encoder network and decoder network on MUG facial dataset. We adopt ResBlock down and up from \cite{brock2018large}. The dimensions of $\vz^c$, $\vz^m_t$, $\vh_t$, $\va$ are $150$, $16$, $180$ and $6$ respectively. The batch size on MUG facial dataset are $30$ and the length of video sequence for training is $T=8.$\label{fig:paramreal}} 
\end{figure}

% \subsection*{Appendix H: Additional Results on TIMIT Dataset}
\subsection*{Appendix H: Additional Results on Audio Data}
\begin{figure}
\centering
\includegraphics[width=0.4\textwidth]{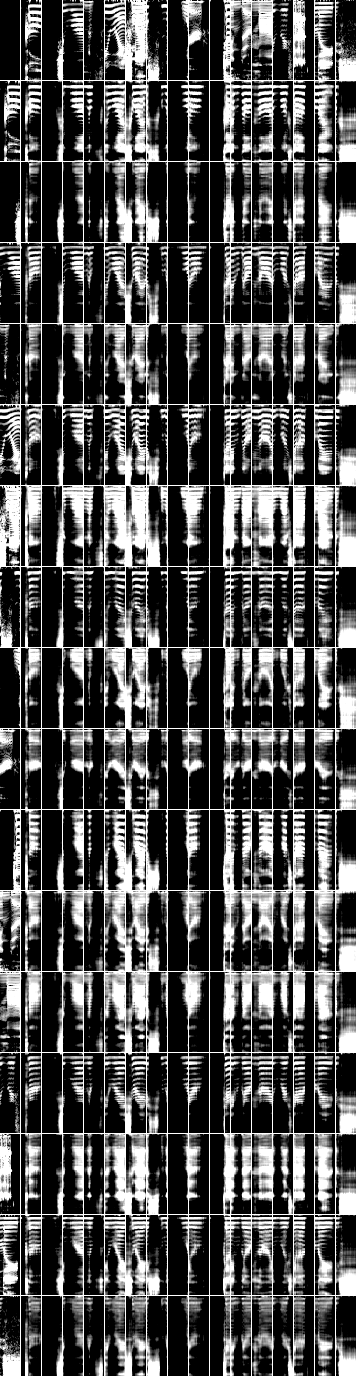}
\caption{Cross generation of 16 audio clips forms a $17\times 17$ matrix. The first column and the first row are spectrum visualization of the original sequences. Subplot at the $(i+1)$-th row and $(j+1)$-th column represents the reconstruction of $i$-th static factor and $j$-th dynamic factor.}
\label{fig:timit_matrix}
\end{figure}

\paragraph{Swapping Static and Dynamic Factors on Audio Data}  Here we present results of swapping static and dynamic factors of given audio sequences. Results are given in Figure~\ref{fig:timit_matrix}. Each heatmap subplot is of dimension $80\times20$ and visualizes the spectrum of 200ms of an audio clip, in which the mel-scale filter bank features are plotted in the frequency domain (x-axis represents temporal domain with 20 timesteps and y-axis is the value of frequencies). We collect these heatmaps in a matrix where the static factors in a row are kept the same and each column shares the same dynamic factor. It can be observed that in each column, the linguistic phonetic contents as reflected by the formants along x-axis are kept almost the same after swapping. Likewise, the timbres are reflected as the harmonics in the spectrum plot. This can be concluded by observing that the horizontal light stripes which represents the harmonics are kept consistent in a row.
Moreover, we perform identity verification experiment as conducted in DS-VAE~\citep{li2018disentangled}. Similar to cross reconstruction, $z^c_\text{female}$ and $z^c_\text{male}$ (or $f^\text{female}$ and $f^\text{male}$ in DS-VAE) are swapped for two sequences $\{x_\text{female}\}$ and $\{x_\text{male}\}$. By an informal listening test of the original-swapped speech sequence pairs, we confirm that the speech content is preserved and identity is transferred (i.e. female voice usually has higher frequency).
%\newpage
\subsection*{Appendix I: Additional Results on a Moving-Shape Video Data}
\paragraph{Generation Results on Moving Shapes}
\begin{table}[h!]
    \centering
    \begin{tabular}{l|cc}\hline
                    & Static Factor Pred. Acc.  & Dynamic Factor Pred. Acc. \\ \hline
    DS-VAE (TFGAN)  & 77.47\% & 72.45\% \\
    DS-VAE (BigGAN) & 75.37\% & 70.85\% \\
    R-WAE (TFGAN)   & 80.50\% & 83.60\% \\
    R-WAE (BigGAN)  & 75.27\% & 80.00\% \\ \hline
    \end{tabular}
    \caption{Prediction accuracy on generated video data, the experiment setting here is similar to Table 2 in the main text. For predicting the static factor, we fix the static latent representation $z^c$ and randomly sample $z^m$, and examine whether the static information is preserved in the generated video (if so, the static attributes should be correctly predicted by a pretrained video classifier). For predicting the dynamic factor, we perform corresponding experiments analogously.}
    \label{tab:my_label}
\end{table}
\begin{figure}[h]
    \centering
    \includegraphics[width=0.95\textwidth]{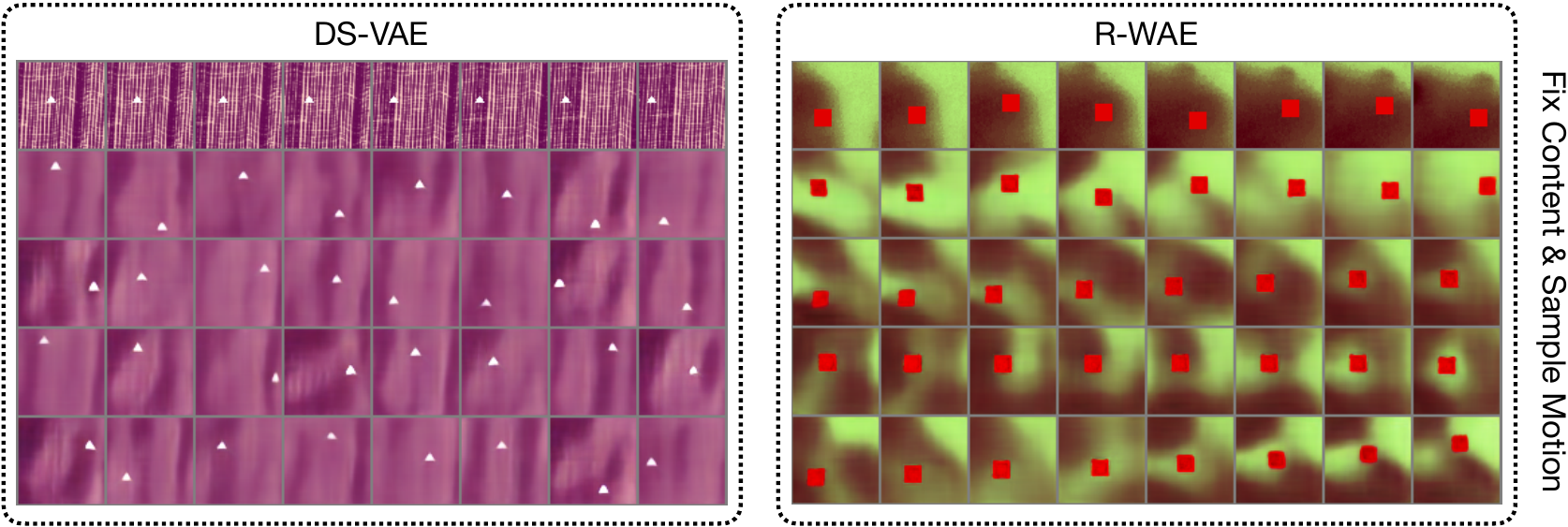}
    \caption{Results of fix $z^c$ and sample $z^m$ using TFGAN~\citep{balaji2018tfgan} architectures. The first row in each subfigure are real video sequences. The generated motion of moving objects by DS-VAE contains abrupt jumps and is not smooth, while R-WAE is able to generate motion of various types including zig-zag, diagonal and straight line. \label{fig:shapes}}
\end{figure}
We report results on a Moving-Shape dataset in Table~\ref{tab:my_label} and Fig.~\ref{fig:shapes}. The Moving-Shape synthetic dataset was introduced in ~\cite{balaji2018tfgan} which has 5 control parameters: shape type (e.g. triangle and square), size (small and large), color (e.g. white and red), motion type (e.g. zig-zag, straight line and diagonal) and motion direction. In Table~\ref{tab:my_label}, TFGAN~\citep{balaji2018tfgan} encoder and decoder architectures are considered less expressive compared with BigGAN~\citep{brock2018large} architectures. Similar to results in Table 2, with more complex and expressive architecture, learning disentangled representation is harder. The results in Table~\ref{tab:my_label} and Fig.~\ref{fig:shapes} demonstrate that R-WAE produces better disentanglement and generation performance than DS-VAE both quantitatively and qualitatively. Qualitative difference of fixing $z^m$ and sampling $z^c$ for DS-VAE and R-WAE is not that obvious and thus not shown.
\end{document}